\documentclass[Afour,sageh,times]{sagej}
\usepackage{moreverb,url}

\usepackage[colorlinks,bookmarksopen,bookmarksnumbered,citecolor=red,urlcolor=red]{hyperref}

\newcommand\BibTeX{{\rmfamily B\kern-.05em \textsc{i\kern-.025em b}\kern-.08em
T\kern-.1667em\lower.7ex\hbox{E}\kern-.125emX}}

\def\volumeyear{2016}

\usepackage{caption}
\usepackage{threeparttablex}
\usepackage{multirow}
\usepackage{makecell}

\usepackage{etoolbox}
\makeatletter
\patchcmd{\@makecaption}
  {\scshape}
  {}
  {}
  {}
\makeatletter
\patchcmd{\@makecaption}
  {\\}
  {.\ }
  {}
  {}
\makeatother

\usepackage{todonotes}

\newcommand{\algname}{Lan-grasp}
\newcommand{\graspit}{GraspIt!}

\title{\LARGE \bf
\algname: Using Large Language Models \\ for Semantic Object Grasping\\ and Placement
}

\runninghead{LAN-GRASP}

\title{\algname: Using Large Language Models for Semantic Object Grasping\\ and Placement}

\author{%
Reihaneh Mirjalili\textsuperscript{*}\affilnum{1}, 
Michael Krawez\textsuperscript{*}\affilnum{1},
Yannik Blei\affilnum{1},
Simone Silenzi\affilnum{2},
Florian Walter\affilnum{1, 3}, and
Wolfram Burgard\affilnum{1}
}

\affiliation{\affilnum{*}These authors have equally contributed to this work.\
\affilnum{1}Department of Computer Science and Artificial Intelligence, University of Technology Nuremberg, Germany\
\affilnum{2}Department of Engineering ``Enzo Ferrari'' (DIEF), University of Modena and Reggio Emilia, Italy\
\affilnum{3}Technical University of Munich, Germany}

\corrauth{Reihaneh Mirjalili, Department of Computer Science and Artificial Intelligence,
University of Technology Nuremberg,
Dr.-Luise-Herzberg-Straße 4,
90461 Nuremberg, Germany.}

\email{reihaneh.mirjalili@utn.de}

\usepackage{graphicx}
\usepackage[export]{adjustbox}
\usepackage{amsmath}
\usepackage[capitalise,nameinlink]{cleveref}

\usepackage[T1]{fontenc}
\usepackage{array}
\usepackage{booktabs}

\usepackage{pbox}
\begin{document}
\begin{abstract}
In this paper, we propose \algname{}, a novel approach towards more appropriate semantic grasping and placing. We leverage foundation models to equip the robot with a semantic understanding of object geometry, enabling it to identify the right place to grasp, which parts to avoid, and the natural pose for placement. This is an important contribution to grasping and utilizing objects in a more meaningful and safe manner. 
We leverage a combination of a Large Language Model, a Vision-Language Model, and a traditional grasp planner to generate grasps that demonstrate a deeper semantic understanding of the objects.
Building on foundation models provides us with a zero-shot grasp method that can handle a wide range of objects without requiring further training or fine-tuning. 
We also propose a method for safely putting down a grasped object. The core idea is to rotate the object upright utilizing a pretrained generative model and the reasoning capabilities of a VLM.
We evaluate our method in real-world experiments on a custom object dataset and present the results of a survey that asks participants to choose an object part appropriate for grasping. 
The results show that the grasps generated by our method are consistently ranked higher by the participants than those generated by a conventional grasping planner and a recent semantic grasping approach. In addition, we propose a Visual Chain-of-Thought feedback loop to assess grasp feasibility in complex scenarios. This mechanism enables dynamic reasoning and generates alternative grasp strategies when needed, ensuring safer and more effective grasping outcomes.

\end{abstract}


\maketitle

\section{Introduction}
Objects found in household environments often require a specific way of interaction. For artificial objects, such as tools, the deployment mode can be implied by their design, which ensures functionality and user safety. For instance, a knife should typically be held by the grip, not the blade. Similarly, a mug with hot tea is best held by the handle and not the rim of the mug. Incorrect handling can also impair the object itself, e.g., trying to carry a plant by the leaves would most likely lead to damage. Through experience, humans develop an intuitive understanding of objects, their parts, and their proper usage. As robots become increasingly integrated into human living environments, it is crucial to equip them with the same kind of semantic knowledge. 

\begin{figure}[ht!]
\centering
\includegraphics[clip,trim=3.5cm 4cm 3.7cm 3cm,width=0.9\linewidth]{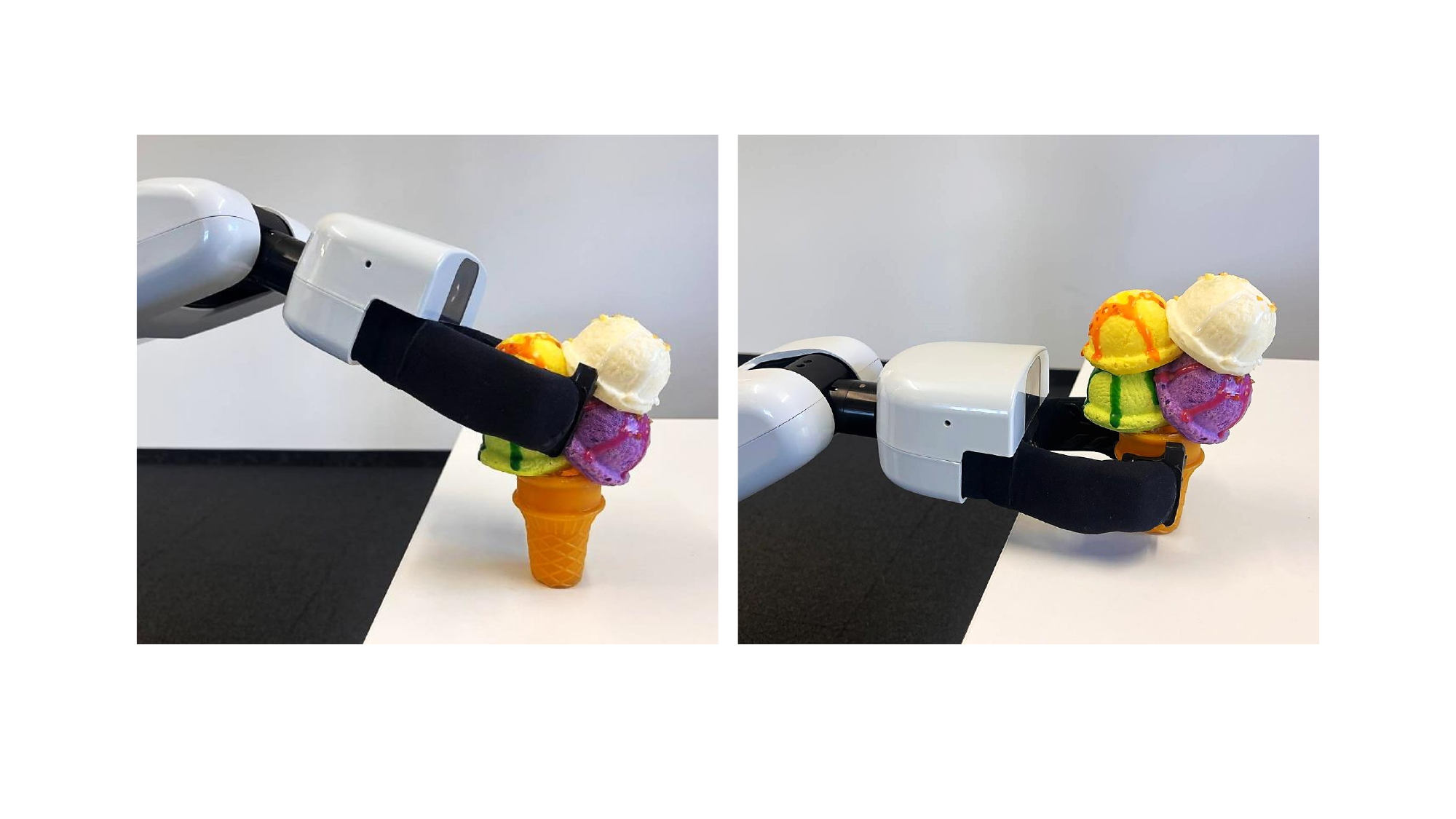}  
\caption{Robot performing the command of \emph{``Pick up the ice cream please''}. The grasp on the left is generated without including semantic information, while the grasp on the right is performed using our method, which leverages a deeper understanding of the task and the object provided by Large Language Models.}
\label{fig:coverpic}
\end{figure}

Traditional approaches to robotic grasping~\citep{bicchi1995closure,miller2000graspit,ten2017grasp} only analyze the object geometry and aim to optimize the grasp stability. Without a deeper understanding of semantic aspects as described above, this can limit the usability of tools or result in object or robot impairment. Recent data-driven approaches~\citep{jang2017end,kwak2022semantic,wu2023learning} also account for the object class and can generate grasps appropriate for the specific object type. Several works tackle the problem of task-specific grasping, where the object is grasped differently depending on the action at hand ~\citep{murali2021same,tang2023task,tang2023graspgpt}. However, most of these methods require substantial computational resources for training and can fail to generalize to unseen object categories. Our objective is to develop an approach for object-specific grasping that ensures tool usability and safety without requiring any training.

We proceed towards this goal by introducing \algname{}, a zero-shot method built on foundation models. The scale of these models and the massive size and generality of their training data allow us to reason about a large variety of objects without further training or fine-tuning. In particular, \algname{} uses a Large Language Model (LLM) to understand which part of an object is suitable for grasping. Next, this information is grounded in the object image by leveraging a Vision-Language Model (VLM). Our method uses GPT-4 as the LLM and OWL-ViT~\citep{owl-vit} as the VLM. However, due to the modular structure of \algname{}, it can be easily adapted to utilize other LLMs or VLMs. Finally, we use an off-the-shelf grasp proposal tool~\citep{miller2000graspit} to plan the grasps in accordance with the admissible parts of the object detected by the deployed foundation models.

We further propose an approach for putting down a grasped object in a safe manner. So far, object placing methods have been mostly focused on placement stability. However, this is not sufficient for safely handling many household items, such as cups and open bottles, which have a defined upright orientation. We argue that placing objects upright not only addresses safety issues but also tackles placement stability, as most such objects are per design stable in an upright position. We deploy the SAM 3D~\citep{chen2025sam} model to generate a 3D mesh of the grasped object and retrieve the object's pose in the input image. We then render the mesh from different views and use a VLM to select the view showing the object upright. 

\textbf{In summary, we make the following contributions: }
\begin{enumerate}
    \item We propose a novel approach using foundation models for zero-shot semantic object grasping.
    \item We demonstrate that the presented approach can work with a wide variety of day-to-day objects without requiring additional training.
    \item We evaluate our approach by asking human participants to choose the appropriate grasps.
    \item We employ a feedback mechanism using Visual Chain-of-Thought prompting to assess grasp feasibility and dynamically generate alternative grasp strategies when needed.
    \item We propose an object placement approach that finds upright object orientations in zero-shot manner using a generative 3D foundation model and a VLM.
\end{enumerate}


\section{Related Work}
\label{sec:related_work}

Traditional grasping algorithms~\citep{bicchi1995closure,miller2000graspit,ten2017grasp,zapata2019fast,myers2015affordance} analyze the geometry of the object and the gripper to propose and evaluate a grasping pose. Building on decades of development, these methods are fast and reliable off-the-shelf tools. However, they do not incorporate semantic information and operate based on object shape only. Also, such methods rely on a precise object model and thus suffer from partial or noisy geometry. Data-driven approaches regress grasping candidates from either single-view RGB images~\citep{jang2017end,redmon2015real} or point clouds~\citep{zhao2021regnet,alliegro2022end}, thus mitigating the need for a complete object model. Further, a network can learn a more natural grasping policy if human-like grasps are included in the training data, where such grasps are either created manually~\citep{corona2020ganhand} or learned through imitation~\citep{wu2023learning}.

Our work is closely related to task-oriented grasping (TOG) and affordance detection. TOG methods restrict the grasp candidates to a specific object part or area, conditioned on the action at hand. \cite{murali2021same} create a data set with a large number of objects and tasks and manually annotate task-specific grasp poses. Then, the authors use that data to train a grasp evaluation network. \cite{kwak2022semantic} deploy a knowledge graph to select the gripper type and gripping force appropriate for the given object. \cite{chen2021joint} propose a network that jointly detects an object and generates a grasping pose according to a natural language command. However, the training requires object, grasp, and command ground truth data. \cite{fang2020learning} introduce TOG-Net, which optimizes task-oriented grasps and manipulation policies using simulated self-supervision.

Similar to TOG, affordance detection is the problem of identifying objects or object parts that accommodate a certain action. \cite{do2018affordancenet} propose an end-to-end trained network that detects object instances in an image and assigns pixel-wise affordance masks to object parts. \cite{liu2020cage} build on the previous work as a backbone for affordance detection and, in addition, infer the material of object parts to further facilitate semantic grasping. \cite{monica2020point} propose a system that decomposes an object point cloud into meaningful parts which then serve as grasping targets. However, the part the robot has to grasp is provided by the user, whereas in our method, the part is suggested by an LLM. \cite{bbohg2013data} survey data-driven approaches to grasp synthesis, focusing on methods that sample and rank candidate grasps for both familiar and unknown objects, highlighting the role of feature extraction in these approaches. \cite{nasiriany2024pivot} introduce a prompting framework for VLMs (PIVOT) that refines candidate actions iteratively, demonstrating potential for spatial tasks such as grasping. But their focus is broader, addressing both navigation and manipulation tasks. \cite{wei2024grasp} propose a novel dataset, DexGYSNet, and utilize it to train a model for dexterous grasp generation based on language guidance. \cite{jian2023affordpose} introduce AffordPose, a large-scale dataset for affordance-driven hand-object interactions, focusing on part-level affordance labeling to guide the generation of hand-object interactions in fine detail. \cite{zhu2021toward} propose a framework for human-like dexterous grasping, using semantic touch codes and object functional areas to guide grasps. \cite{ren2023leveraging} introduce ATLA, a meta-learning framework that uses LLMs to accelerate tool manipulation by combining language-based policies with affordances, focusing on general tool use rather than grasping specific object parts.

Foundation models have recently attracted a lot of attention in different sub-fields of robotics~\citep{mirjalili2023fm-loc,huang2023audio,cui2023no} and
have also been applied to boost TOG and affordance detection. \cite{ngyen2023open} train an open-vocabulary affordance detector for point clouds whereby CLIP is deployed to encode the affordance labels. Similarly, \cite{tang2023task} use CLIP to facilitate task-specific grasping from RGB images and language instructions. The authors propose utilizing CLIP embeddings from intermediate CLIP layers to enable their affordance detector to reason about fine-grained object parts. \cite{gao2023physically} annotate a large object data set with physical object properties like mass or fragility and fine-tune a VLM on it to improve manipulation planning. Other methods integrate LLMs for encoding tasks or object parts from natural language. \cite{song2023learning} use BERT as the language back-end and train a network that grounds object parts in a point cloud from a user instruction. Here, however, the part label is explicitly referred to in the user input. The approach of \cite{tang2023graspgpt} lifts this limitation by prompting an LLM to describe the shape and parts of an object. The LLM response is then processed by a Transformer-based grasp evaluation network. Our method also relies on an LLM to decide which object part should be grasped. The crucial difference to the above works is that our approach relies solely on foundation models and does not require any training. Thus, once more powerful foundation models are available, the performance of our approach is easily improved by switching to a new LLM or VLM. \cite{newbury2023deep} conduct a systematic review of deep learning methods for six-DoF grasp synthesis, highlighting sampling-based, direct regression, and reinforcement learning methods to generate grasp poses. \cite{wu2024see} propose an approach to enhance the robustness of large multimodal models in vision tasks by introducing reasoning capabilities to correct false premises, which improves reasoning for affordance-based grasping tasks. In a similar line of thought, \cite{huang2023voxposer} propose VoxPoser, a framework that generates 3D value maps to guide robotic manipulation using affordances extracted from LLMs and visual grounding. However, their focus is on manipulation tasks rather than detecting the specific grasping part of an object.

Finally, embodied VLMs like PALM-E~\citep{driess2023palm} aim to close the gap between language, vision, and robot actions by jointly training the network on these modalities. Vision-Language Action Models~\citep{rt1,brohan2023rt,openx,octo,pizero,openvla} extend the idea of foundation models to robotics by training or fine-tuning large models, typically based on the Transformer architecture, on large-scale robotics datasets~\citep{openx}. They commonly employ a behavioral cloning objective to learn from human demonstrations and corresponding language instructions. A language instruction, thus, results in the execution of the corresponding skill. As the training of these models is based on demonstrations, they can be expected to implicitly align with human preferences for grasping within the domains of the training datasets. In contrast, our method explicitly aligns with human preferences, supports grasp feasibility feedback, and is not limited to a specific dataset.

The problem of object placement in robotics has been mainly approached from two directions. The first is considering the object geometry and physical properties to achieve a stable object position after placement. The second focuses on human-like object arrangement. 
\cite{baumgartl2014fast} develop a geometry-based placement planner which finds a stable position for arbitrary objects on uneven surfaces.  
\cite{haustein2019object} tackle motion planning for placing household items in cluttered environments. Their method considers object stability constraints and human preferences for placing objects next to each other. More recent works integrate learning-based methods and pretrained foundation models to improve object placement. The approach of \cite{lee2024spots} combines a simulation-based stability verification with a contextual placement area selection driven by an LLM. \cite{mendez2024everyday} deploy a diffusion model to generate a human-like object arrangement on a 2D surface and use this layout to guide object positioning. A diffusion model is also used by \cite{nadeau2025generating} to generate physically stable object placements. These methods, however, do not consider the placement orientation of objects. The notion of \textit{upright} or \textit{canonical} object orientation has been explored in the computer vision and computer graphics communities~\citep{fu2008upright,wang2025orient,lu2025orientation}. In robotics, \cite{newbury2021learning} propose a two-part approach for placing an object upright. They train one CNN that regresses the angle for rotating the object upright, and one verification network that judges the current orientation. \cite{pang2022upright} also propose a data-driven method for finding the upright orientation of object point clouds, where a CNN first identifies the support base plane and the upright orientation is computed from the plane normal. These models are trained on custom datasets with potential training data limitations, whereas we rely on large pretrained models like ChatGPT and SAM 3D~\citep{chen2025sam}.

\section{Method Description}

\begin{figure*}[ht]
\centering
\includegraphics[clip,trim=0cm 3.5cm 0cm 2cm,width=1\linewidth]{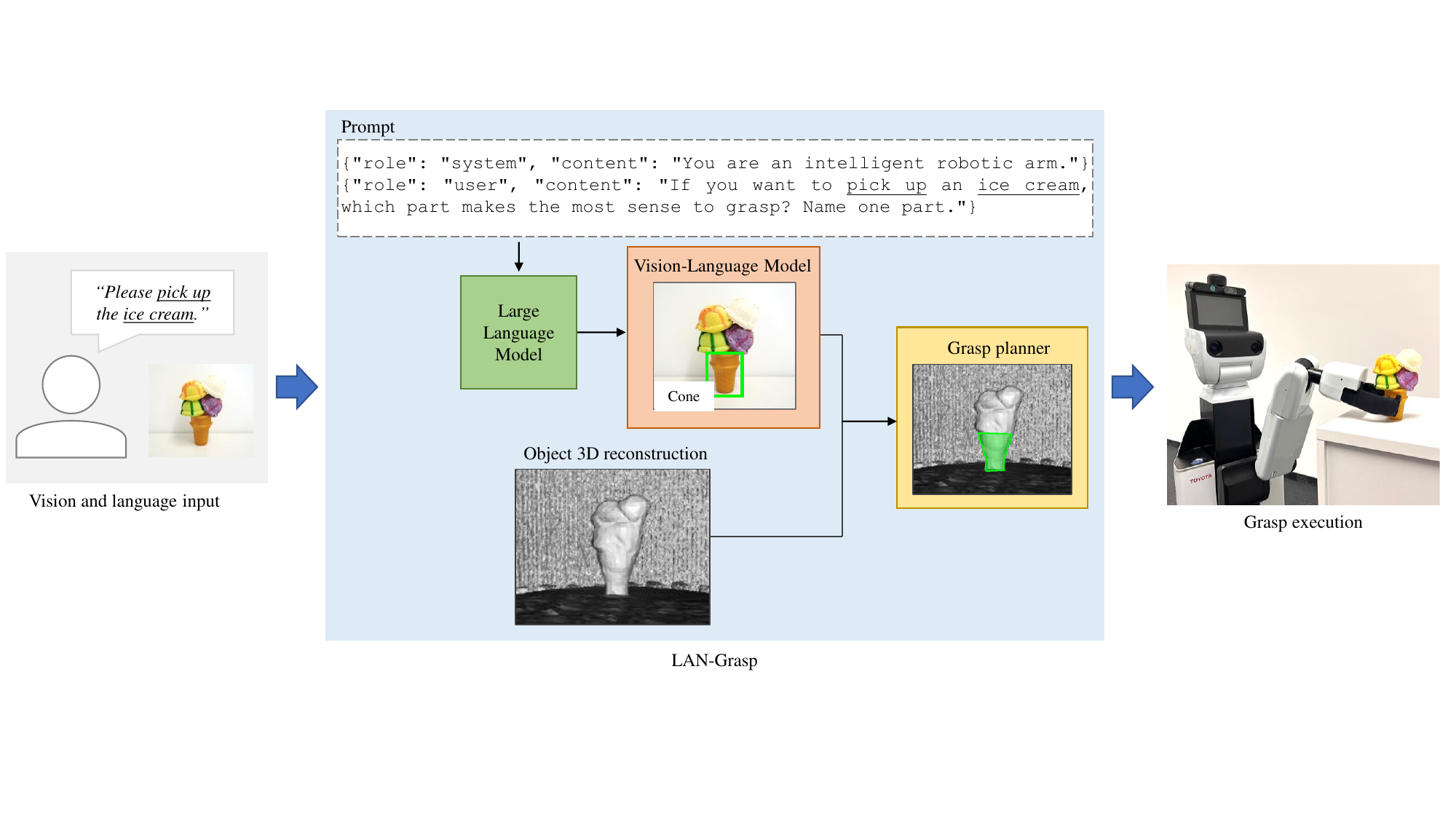}  
\caption{Our grasping approach in a nutshell: The command from the user is turned into a prompt suitable for the Large Language Model (LLM). With this prompt as an input, the LLM outputs the proper part for grasping the object, which in this example is the \emph{cone}. This word is then grounded to the object image using a Vision-Language Model (VLM). The grounded grasp part is integrated into the 3D reconstruction model of the object to generate the proper grasp.}
\label{fig:overview}
\end{figure*}

In this section, we explain the details of our method. \algname{} generates a grasping pose from an object label, a camera image showing that object, and the corresponding object geometry. The method consists of two parts. In the language module, an LLM first decides what object part to grasp, which is then grounded in the image by a VLM. The resulting bounding box is projected onto the object geometry to mark the grasp target in the grasp planning module. Thanks to the modular structure of \algname{}, it is easy to enhance the pipeline by leveraging more advanced models as they emerge. The pipeline of \algname{} is depicted in \cref{fig:overview}.

\subsection{Language Module}
\label{sec:method_language}
In the first step, the pipeline generates an LLM prompt from the object label~{\tt<object>} provided by the user. The scheme of the prompt is chosen to be compatible with GPT-4, the LLM we use in the pipeline~\citep{openai2023gpt4}. \cref{fig:overview} contains an example for an ice cream cone. We included the last sentence in the prompt to prevent the LLM from giving extra explanations and thus only output the desired object part.
We use OWL-ViT~\citep{owl-vit} as the VLM for grounding the object part label in the image. 
It builds on the Vision Transformer Architecture, first presented by \cite{dosovitskiy2020image}. 
The authors first pre-train the model using contrastive learning \citep{zhai2022lit} on a large image-text data set \citep{jia2021scaling}.
Afterward, they fine-tune it on publicly available object detection datasets. During inference, OWL-ViT detects and marks the desired object part with a bounding box, which we then project onto the object's 3D model.

\subsection{Grasp Planning Module}
\label{sec:grasp_planning}
We deploy the \graspit{} simulator~\citep{miller2004graspit} as our grasp proposal generator. It is a standard tool that operates on geometric models and evaluates grasps according to physical constraints. Thus, the first step for grasp planning is to create a dense 3D mesh model of the object. In our setup, we use two fixed RGB-D sensors and a turning table for object scanning. We acquire the camera poses from an Aruco board and integrate the depth images via KinectFusion~\citep{newcombe2011kinectfusion}. However, we note that any other suitable reconstruction approach could be used here. 

The possible poses for grasping the object are generated by sampling. The initial gripper position is selected based on the object's geometry. After that, the gripper is iteratively brought closer to the object while avoiding obstacles~\citep{miller2003automatic}. In this regard, \graspit{} splits the scene into object and obstacle geometry, and we exploit this mechanism by marking the mesh parts that project into the VLM-generated bounding box as object and the rest as obstacle. This enforces grasping only at the desired object part. The resulting grasp proposals are ranked based on grasp efficiency and finger friction.

We want to point out that our approach is agnostic about the grasp planner and could be potentially replaced by other tools that do not require a complete object model, e.g., the method of \cite{alliegro2022end}. In this case, the reconstruction step could be skipped entirely, and the grasp candidates could be computed on a point cloud acquired from the robot's sensors.

\subsection{Object Placement}
\label{sec:placement_method}

\begin{figure*}[ht]
\centering
\includegraphics[clip,trim=0cm 0cm 0cm 0cm,width=0.8\linewidth]{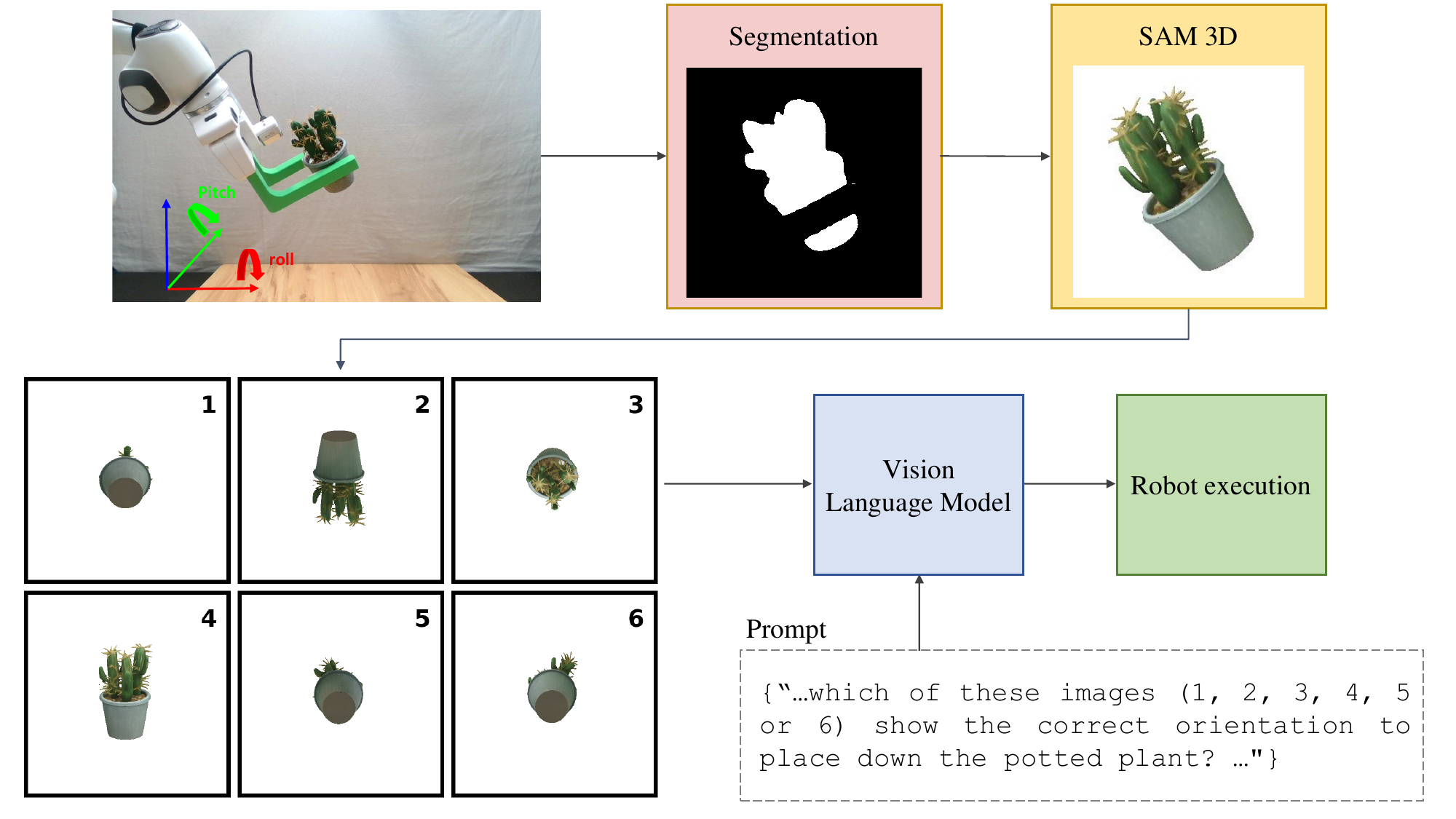}  
\caption{Summary of our placing approach showing one object alignment cycle. Given an RGB image of the grasped object, we first segment the object and then use SAM 3D to obtain the object 3D reconstruction and pose estimation. We then render the reconstructed mesh from six views, each showing one axis in the object frame up or down . A vision–language model is prompted to select the orientation that best matches placing the object down correctly. Finally, the robot executes the corresponding roll, pitch, and yaw rotation to align the grasp to the selected object pose.}
\label{fig:placement_overview}
\end{figure*}

We also address the task of determining the correct orientation of an object for placing it down. For instance, an open bottle should only be placed on its bottom part; putting it on the side or upside-down could lead to liquid spillage. Similar holds for many other household objects like mugs, glasses, plates, etc. However, ultimately, a safe placement orientation requires common-sense knowledge about each individual object. If such an orientation is meaningful for an object, we refer to it as the $up$-orientation in the following. As with grasping, we want to extract the knowledge about placement orientations from large pretrained models in a zero-shot fashion.

In our setup, we assume that the robot gripper holds an object in front of a static camera and needs to rotate the object correctly before placing it on a flat surface. The correct object orientation is determined by a VLM. Preliminary experiments have shown that generalist VLMs can detect when an object is oriented upright in the image, but struggle to reason how it should be rotated if the initial orientation is off. A brute-force solution could be to rotate the object in front of the camera in regular steps until the $up$-orientation is found. Depending on the step size, however, this procedure might take many object adjustment and VLM inference cycles until termination. Therefore, we propose a more direct approach to finding the placement orientation. 

We deploy the recently released SAM 3D model~\citep{chen2025sam} that generates a 3D mesh $M$ from an image and an object mask. It also estimates the transform $T_{o2c} \in SE(3)$ from the object coordinates $o$ to the camera frame. We can now render the object mesh from arbitrary view points $v \in V$, select the view $up \in V$ with the object upright via the VLM, and compute the necessary gripper rotation from $T_{o2up} T_{o2c}^{-1}$. The most general but also most inefficient strategy to construct $V$ is to sample the views uniformly from a sphere around the object. Cost-reducing heuristics could be to first compute the 3D bounding box around $M$ or the principal components of points in $M$ and assume that the object's $up$-direction is aligned with one of the axes, thus reducing the number of views to $|V|=6$.

However, we also observe that the object frame $o$ generated by SAM 3D already fulfills this assumption, i.e., the $up$ mesh orientation is aligned with one of the axes in $o$, with the choice of the $up$-axis depending on the initial object pose in the image. We note that this is not an officially documented feature of SAM 3D, but was persistent for all objects in our experiments. We speculate that it is a byproduct of training the model on generic object mesh datasets, where the canonical mesh frame is typically upright. Further details and a comparison with the bounding box and principal component heuristics are described in \cref{sec:placement_evaluation}.

\begin{figure*}[ht]
\centering
\includegraphics[clip,trim=0cm 12cm 0cm 0cm,width=1\linewidth]{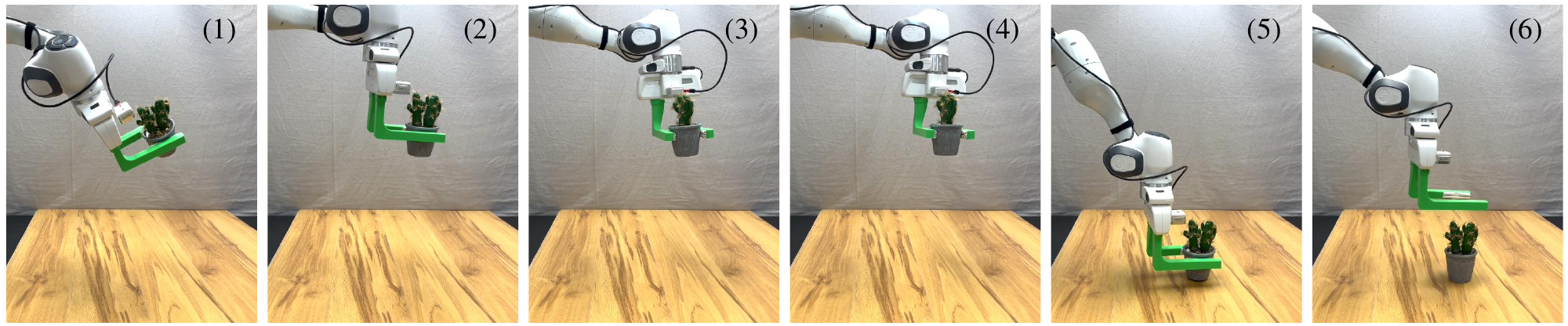}  
\caption{Example sequence from a single placing trial: (1) initial camera image with the grasped object, (2) the estimated pose from the first alignment cycle is applied, (3) a 90° yaw rotation reorients the grasp to prepare for the second cycle, (4) the estimated pose from the second alignment cycle is applied, (5) the object is placed on the surface, and (6) the gripper opens to release the object.}
\label{fig:placement_sequence}
\end{figure*}

We construct $V$ such that each axis in $o$ is pointing up or down in one view with respect to the object coordination frame, i.e:
\begin{equation*}
V = \{X^{obj}_{up}, X^{obj}_{down}, Y^{obj}_{up}, Y^{obj}_{down}, Z^{obj}_{up}, Z^{obj}_{down} \} 
\end{equation*}
\cref{fig:placement_overview} shows the six resulting renderings. Next, these renderings are sent to the VLM to select the correct placement orientation and subsequently move the robot. We refer to one run of this pipeline as one \textit{alignment cycle}. All rotations are performed in the robot world coordinate frame depicted in \cref{fig:placement_overview}, with the center of rotation at the gripper tool center point (TCP).
The robustness of this method depends on the initial object pose estimate $T_{o2c}$. Our experiments show that SAM 3D is sufficiently reliable for estimating pitch rotations in the robot world frame, but it performs worse for roll rotations, i.e., when the object is tilted away from the camera. Therefore, after the first alignment cycle, we rotate the gripper by a $90^{\circ}$ yaw about the $z$-axis and perform a second alignment cycle to compensate for inaccuracies in the roll-angle prediction from the first cycle. After that, the object is lowered down until a contact with the table surface is detected by robot proprioception. The full alignment sequence is depicted in \cref{fig:placement_sequence}.

\section{Experimental Evaluation}
\label{sec:results}
In this section, we present the details of our experiments and results. Our goal is to demonstrate that our method proposes to grasp object parts that are preferred by humans across a variety of objects. We argue that humans generally choose grasps that enable correct tool usage and ensure safety, and that a robot retains these desirable qualities by executing similar grasps. To that end, we first collect a dataset of typical household objects. Next, we apply our approach to these objects and execute the grasping on a real robot. We show that our grasping strategy is similar to human preferences obtained through a survey and that our approach outperforms two baselines based on this similarity metric. Finally, we evaluate our proposed placement method on a range of different objects.

In the following sections, we describe our dataset, provide details on the baseline approaches and the experiments performed, and discuss the results. In \cref{sec:feasibility_feedback}, we present an extension to the main algorithm that reasons about the feasibility of a grasp in complex scenarios. In \cref{sec:ablation}, we perform an ablation study on several components of the pipeline. Finally, in \cref{sec:placement_evaluation} we present the setup details, results, and discussion for the object placement approach.

\subsection{Dataset}
\label{sec:dataset_original}
We collect a dataset containing $22$ different objects commonly found in household environments. We chose these objects to cover a wide range of situations where semantic knowledge is required for proper grasping. Our first objective was to showcase the grasping of functional objects like tools or kitchen supplies, e.g., \emph{shovel}, \emph{hand brush}, and \emph{knife}. Furthermore, we included delicate objects that might be damaged with an improper grasp, such as \emph{rose}, \emph{cupcake}, and \emph{ice cream}. For other objects, a wrong grasp can cause a dangerous situation, e.g., \emph{candle}. Finally, we included objects where an improper grasp might not necessarily be harmful but is rather unnatural to a human observer, for instance, \emph{doll}, \emph{bag}, and \emph{wine glass}. The objects in the data set are shown in \cref{fig:tile1,fig:tile2,fig:narrow}.

\subsection{Experimental Setup and Baselines}
For $11$ objects from our dataset, listed in \cref{tab:similarity}, we conduct real-world experiments using the Human Support Robot (HSR)~\citep{yamamoto2019development}. We first scan each object and then apply \graspit{} to the resulting 3D model as detailed in \cref{sec:grasp_planning}. From the top-$20$ grasp proposals, we randomly pick one and execute it using the proprietary HSR motion planner. 

Our first baseline is the plain \graspit{} simulator. Here we use the same 3D models as for our approach but do not restrict grasping to the object part selected by the language model. For each object, we evaluate the top-$20$ grasp proposals and carry out two of them on the HSR, as shown in \cref{fig:tile1}.

The second baseline is GraspGPT~\citep{tang2023graspgpt}, a recent approach to task-oriented grasping. This method requires as input an object point cloud and a natural language prompt describing the object, the object class, and the task. We generate the point clouds from the object meshes reconstructed as above and use an object-specific activity as the task label. Again, we retrieve $20$ grasps per object but do not carry them out on the robot.

\subsection{Qualitative Results}
\begin{figure*}[htbp]
\centering

\includegraphics[clip,trim=1cm 8.07cm 0 1cm,,width=0.9\linewidth]{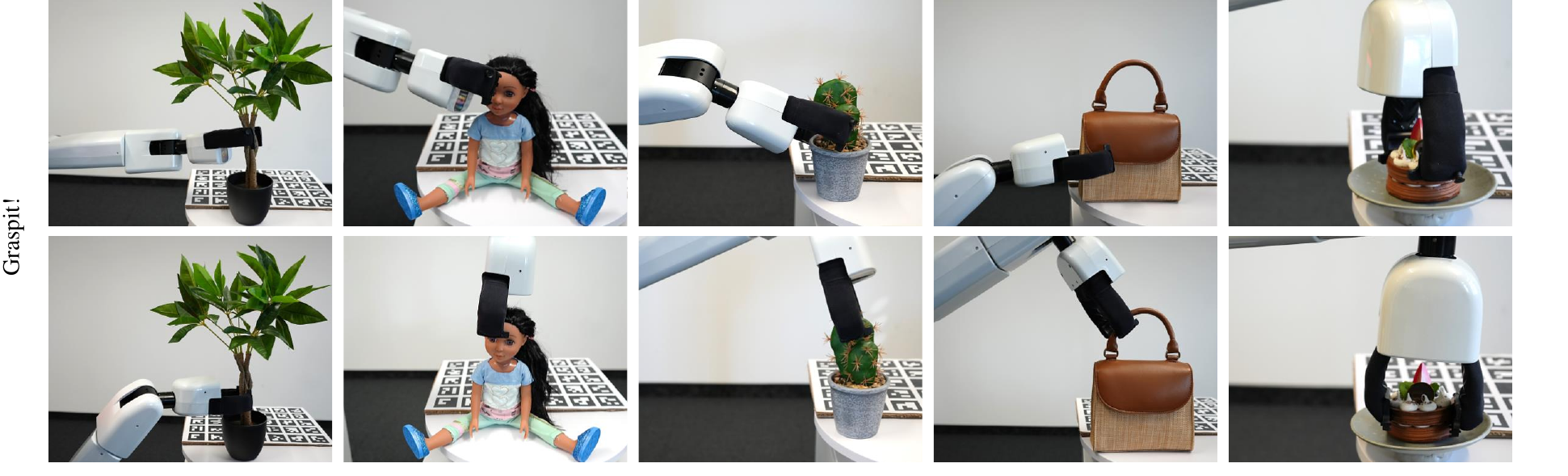}  

\includegraphics[clip,trim=1cm 1cm 0 0cm,,width=0.9\linewidth]{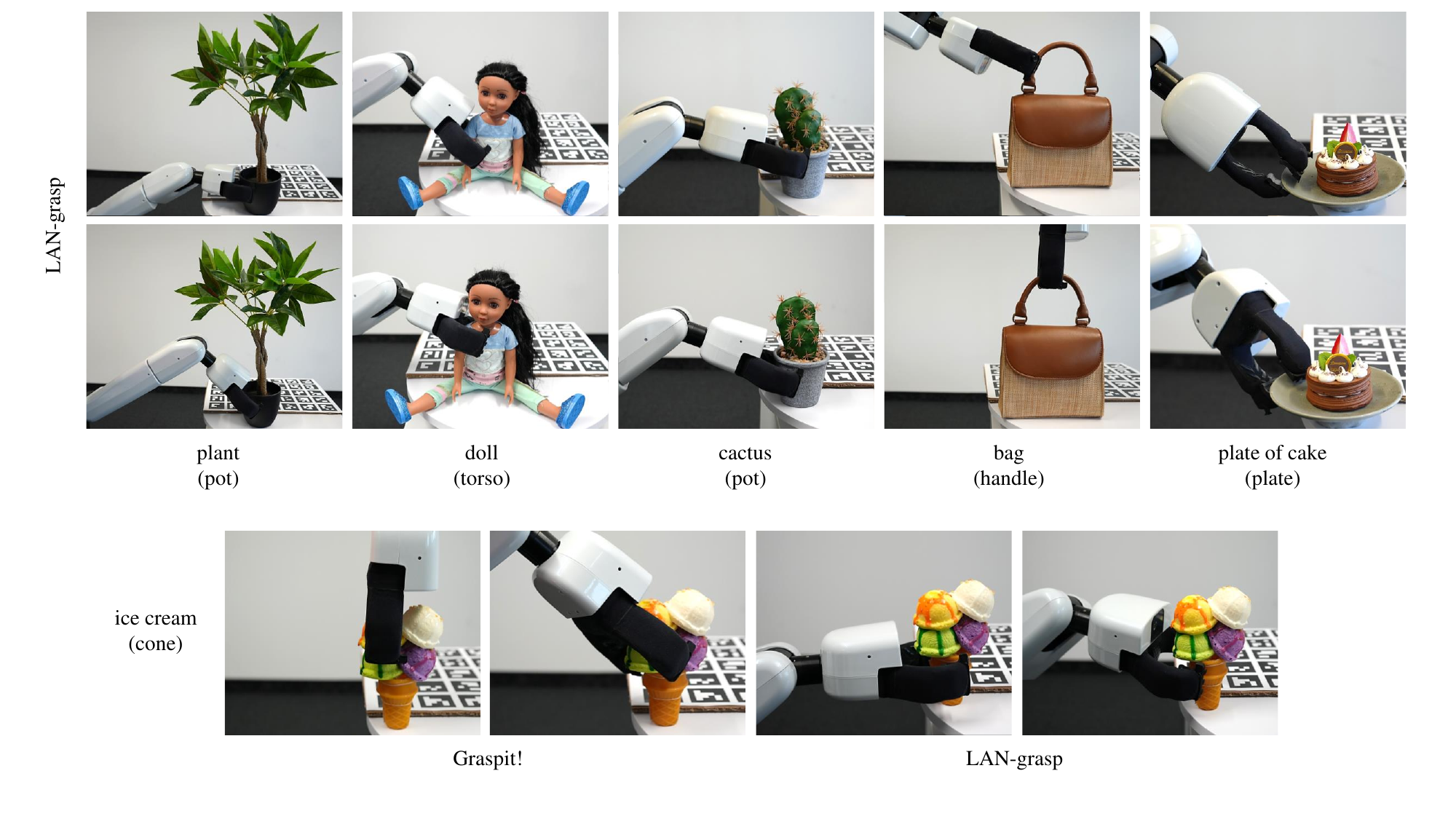}  
\caption{The grasps performed by the HSR robot: Each column presents the grasps for one object. The first two rows for each object show the grasps generated without semantic knowledge about the objects, while the third and fourth rows show the grasps generated by \algname{}.}
\label{fig:tile1}
\end{figure*}

\begin{figure*}[htbp]
\centering
\includegraphics[clip,trim=1cm 8.02cm 0 1cm,width=0.9\linewidth]{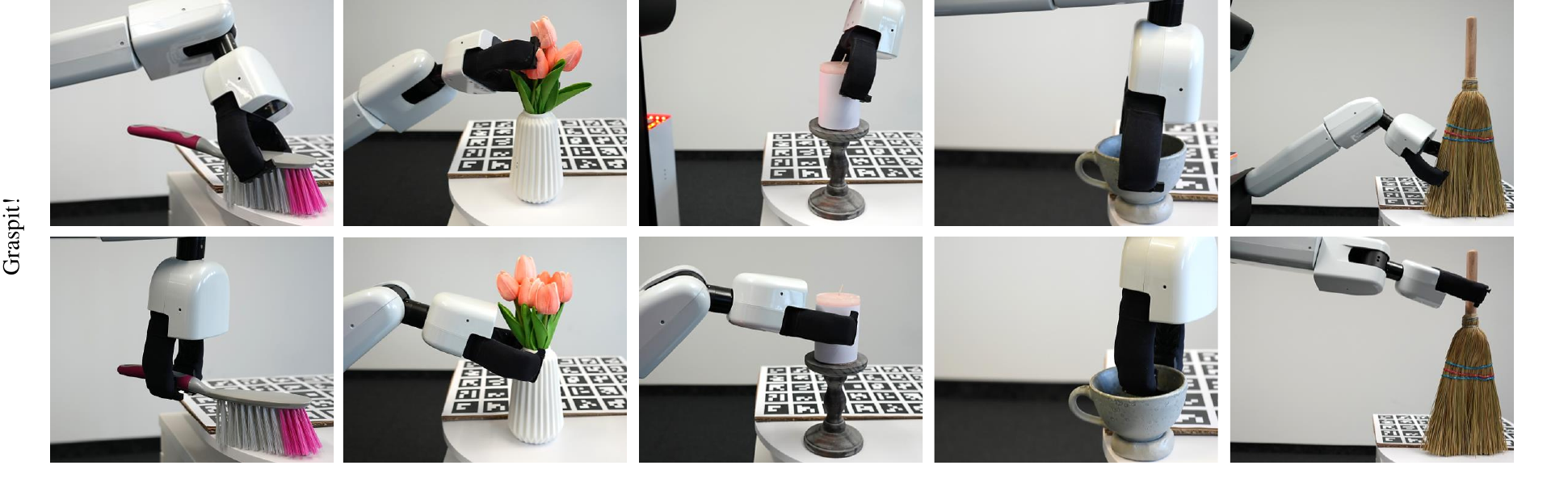}  
\includegraphics[clip,trim=1cm 6.5cm 0 0.7cm,width=0.9\linewidth]{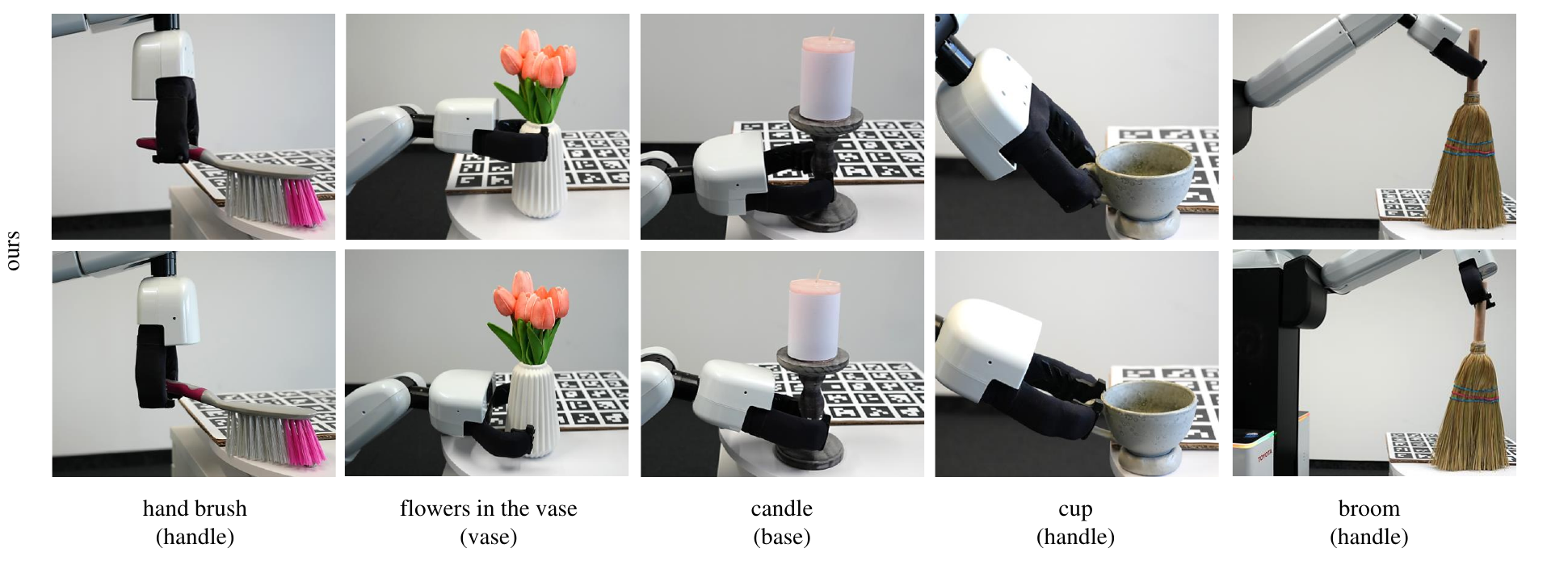}  
\caption{The grasps performed by the HSR robot: Each column presents the grasps for one object. The first two rows for each object, show the grasps generated without semantic knowledge about the objects, while the third and fourth rows show the grasps generated by \algname{}.}
\label{fig:tile2}
\end{figure*}

\begin{figure*}[htbp]
\centering
\includegraphics[clip,trim=1cm 9cm 9cm 0.5cm,width=0.9\linewidth]{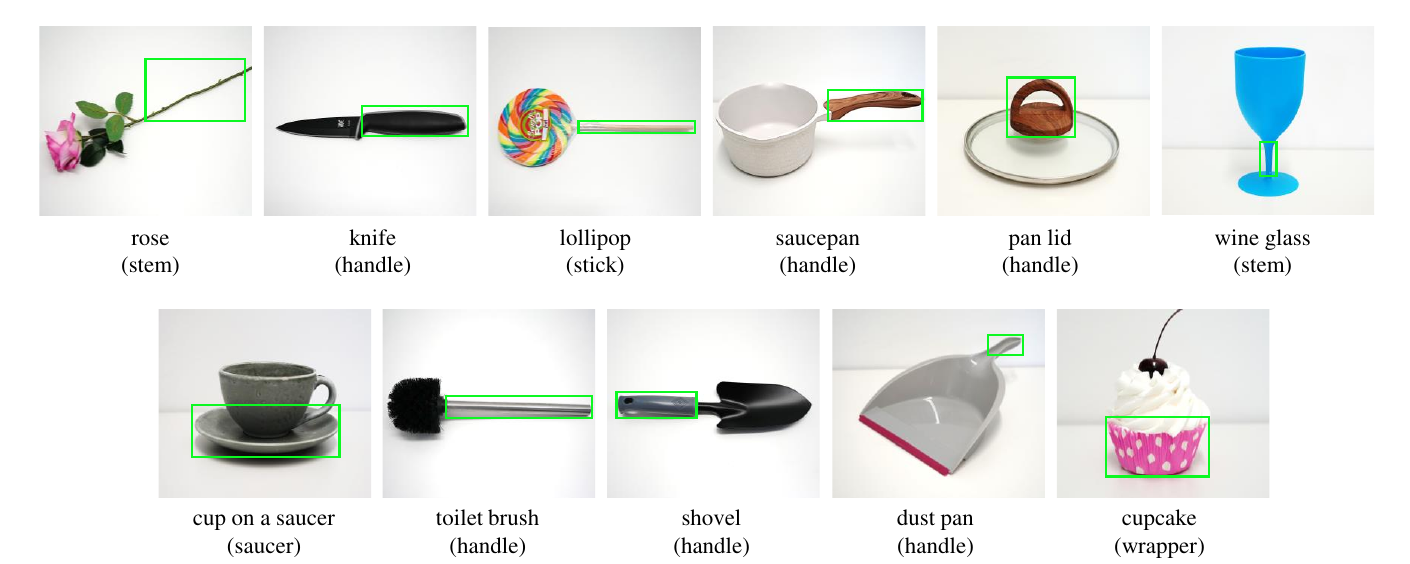}  
\caption{The results of \algname{} on a set of common household objects. The green bounding box shows the area to grasp suggested by the method. }
\label{fig:narrow}
\end{figure*}

In this section, we present and discuss the grasping results of \algname{} and the \graspit-baseline. The grasps executed on the robot are shown in \cref{fig:tile1,fig:tile2}. For the rest of the objects, the grasping area suggested by our method is presented in \cref{fig:narrow}. 

The results suggest that \algname{} proposes grasps suitable for the usage of the respective object.
For instance, grasping the \emph{handle} for \emph{shovel} and \emph{broom} corresponds to the intended use of these items. For \emph{lollipop} and \emph{cupcake}, the grasp is placed away from the edible part at the \emph{stick} and the \emph{wrapper}, respectively. It is noteworthy that our method is able to understand the relation between stacked objects, e.g., ~\emph{flowers in a vase} or \emph{plate of cake}. Also, for a single \emph{cup}, \algname{} suggests grasping the \emph{handle} while for the \emph{cup on a saucer} the grasp proposal is the \emph{saucer}. Other objects, e.g., \emph{doll}, \emph{bag}, or \emph{wine glass}, do not possess a critical area where grasping would cause harm or directly interfere with the functionality. However, our method is able to generate grasps that are closer to how a human would handle these items. In contrast to \algname{}, the areas suggested by \graspit{} are expectantly random and do not consider semantic intricacies.

\begin{table*}[t]
\centering
\footnotesize
\renewcommand{\arraystretch}{1.15}
\setlength{\tabcolsep}{7pt}

\caption[Similarity of grasping area preferences compared to a human user]{Similarity of grasping area preferences compared to a human user. The left half of the table lists the objects and the object part the majority of survey participants suggested for grasping, with the corresponding percentage of users. The right half shows similarity scores for the two baselines and the proposed method.}
\label{tab:similarity}

\begin{tabular}{llccc}
\toprule
\textbf{Object} & \textbf{Preferred Part (\%)} & \textbf{GraspIt!} & \textbf{GraspGPT} & \textbf{Lan-grasp} \\
\midrule
doll                 & torso 92.1       & 0.28 & 0.48 & \textbf{0.92} \\
ice cream            & cone 100.0       & 0.05 & 0.40 & \textbf{1.00} \\
candle               & base 93.1        & 0.22 & 0.57 & \textbf{0.93} \\
flowers in the vase  & vase 93.2        & 0.32 & 0.73 & \textbf{0.93} \\
bag                  & handle 91.1      & 0.79 & 0.69 & \textbf{0.91} \\
plant                & pot 94.3         & 0.16 & 0.56 & \textbf{0.94} \\
hand brush           & handle 95.4      & 0.65 & \textbf{0.95} & \textbf{0.95} \\
toilet brush         & handle 97.6      & 0.42 & 0.52 & \textbf{0.98} \\
cactus               & pot 98.8         & 0.26 & \textbf{0.99} & \textbf{0.99} \\
cupcake              & wrapper 100.0    & 0.10 & 0.40 & \textbf{1.00} \\
cup on a saucer      & saucer 81.2      & 0.24 & 0.59 & \textbf{0.81} \\
plate of cake        & plate 98.8       & 0.11 & 0.51 & \textbf{0.99} \\
mug                  & handle 77.1      & 0.28 & 0.73 & \textbf{0.77} \\
saucepan             & handle 94.3      & 0.36 & \textbf{0.94} & \textbf{0.94} \\
broom                & handle 97.6      & 0.42 & \textbf{0.98} & \textbf{0.98} \\
\midrule
\textbf{Average}     & --               & 0.31 & 0.67 & \textbf{0.94} \\
\bottomrule
\end{tabular}
\end{table*}

\subsection{Quantitative Results}
\label{sec:results_quantitative}
To support the claim that our approach proposes grasps similar to human preferences, we designed a questionnaire on grasping choices. A group of $83$ participants were presented with images of all objects used in the experiments and asked where they would grasp them. For each object, the participants could choose between two parts marked by bounding boxes in the image. The survey results are summarized in \cref{tab:similarity}. For each object, we state the preferred part and the percentage of participants who selected it.

For the proposed approach and the baselines, we aimed to evaluate how similar the generated grasps are to those suggested by human users. Given that an object is segmented into parts $a$ and $b$, let $p_a \in [0,1]$ be the empirical probability that a method grasps at part $a$ and $p_b = 1 - p_a$ that part $b$ is grasped. Further, let $p_a^h$ be the human grasping frequency at $a$ according to the survey results and $p_a^x$ the corresponding frequency produced by one of the considered methods. To compute $p_a^x$ for the baselines, we obtained $20$ grasp proposals from each algorithm and counted the grasps falling into region $a$. \algname{} restricts the grasps to the object part selected by the LLM, which, in our experiments, robustly proposed the same part for a given object. Thus, the values of $p_a^x$ were either $1$ or $0$. Finally, we computed a per-object similarity score for each method $x$ as $sim_x = 1-|p_a^h-p_a^x|$. These scores are shown in \cref{tab:similarity} along with the average similarity scores over all objects.

Our method consistently outperforms the baselines on the similarity score and ties only in four cases with GraspGPT. The average similarity score of \algname{} is considerably higher with the value of $0.94$ compared to $0.31$ achieved by \graspit{} and $0.67$ achieved by GraspGPT. We further note that in all cases, the object part choice of \algname{} coincides with the majority vote of the survey participants. The low score of \graspit{} is not surprising since it only considers geometric and not semantic aspects of the object. Thus, whether the object is grasped in a particular region is purely a matter of chance, and we expect the similarity score to be closer to $0.5$ for a larger dataset. GraspGPT exhibits a better performance compared to \graspit{} due to leveraging semantic concepts and LLMs. However, it was tuned on a dataset mostly containing tools and house supplies and thus does not perform well on objects like a \emph{doll} or an \emph{ice cream}.

\begin{figure*}[ht]
\centering
\includegraphics[clip,trim=0.25cm 3cm 0.25cm 0cm,width=1\linewidth]{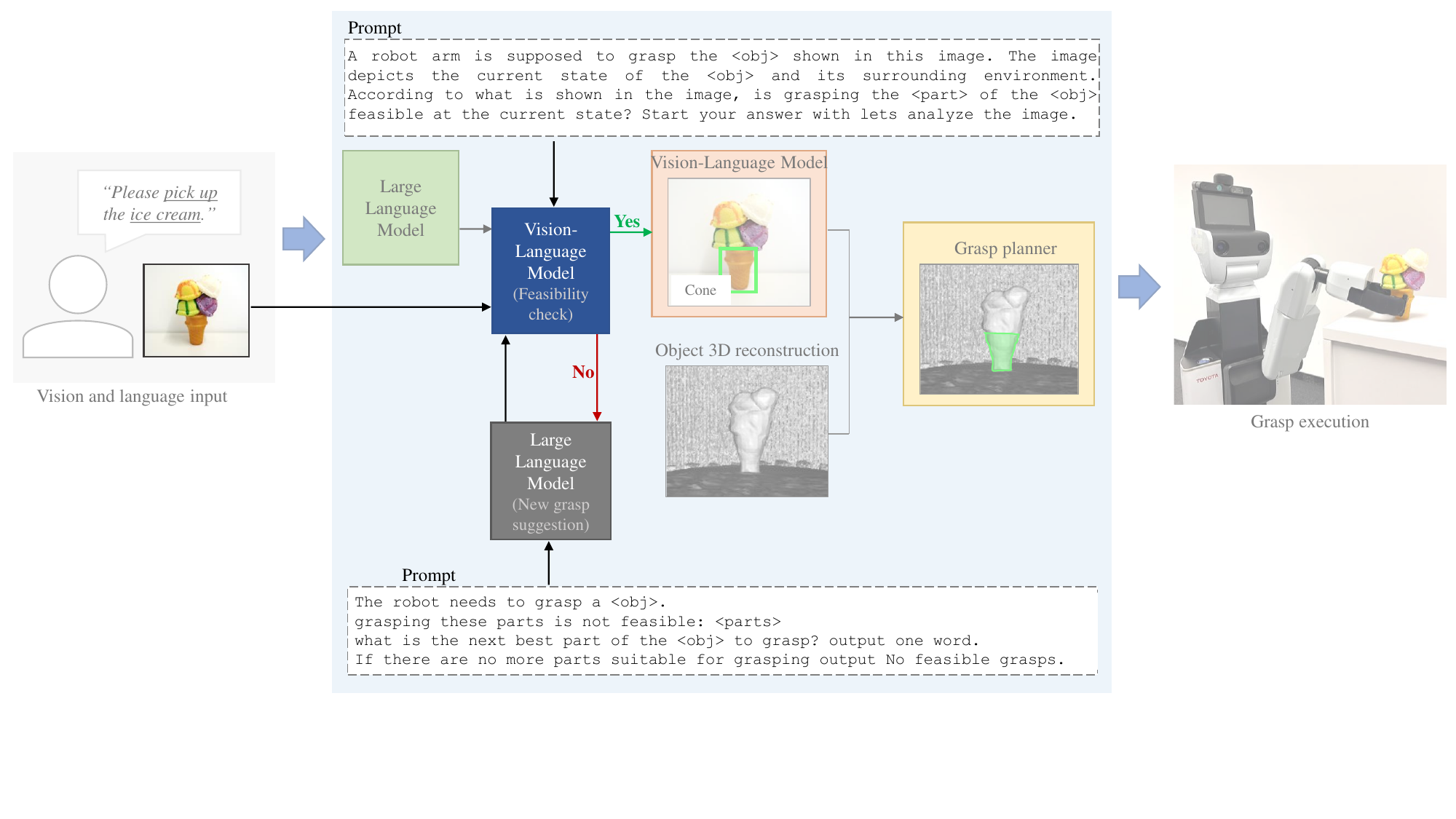}  
\caption{Schematic diagram of the feasibility feedback loop added to the core pipeline of \algname{}.}
\label{fig:main_feedback}
\end{figure*}

\begin{figure*}[ht!]
\centering
\includegraphics[clip,trim=1cm 0.5cm 1cm 0cm,width=0.9\linewidth]{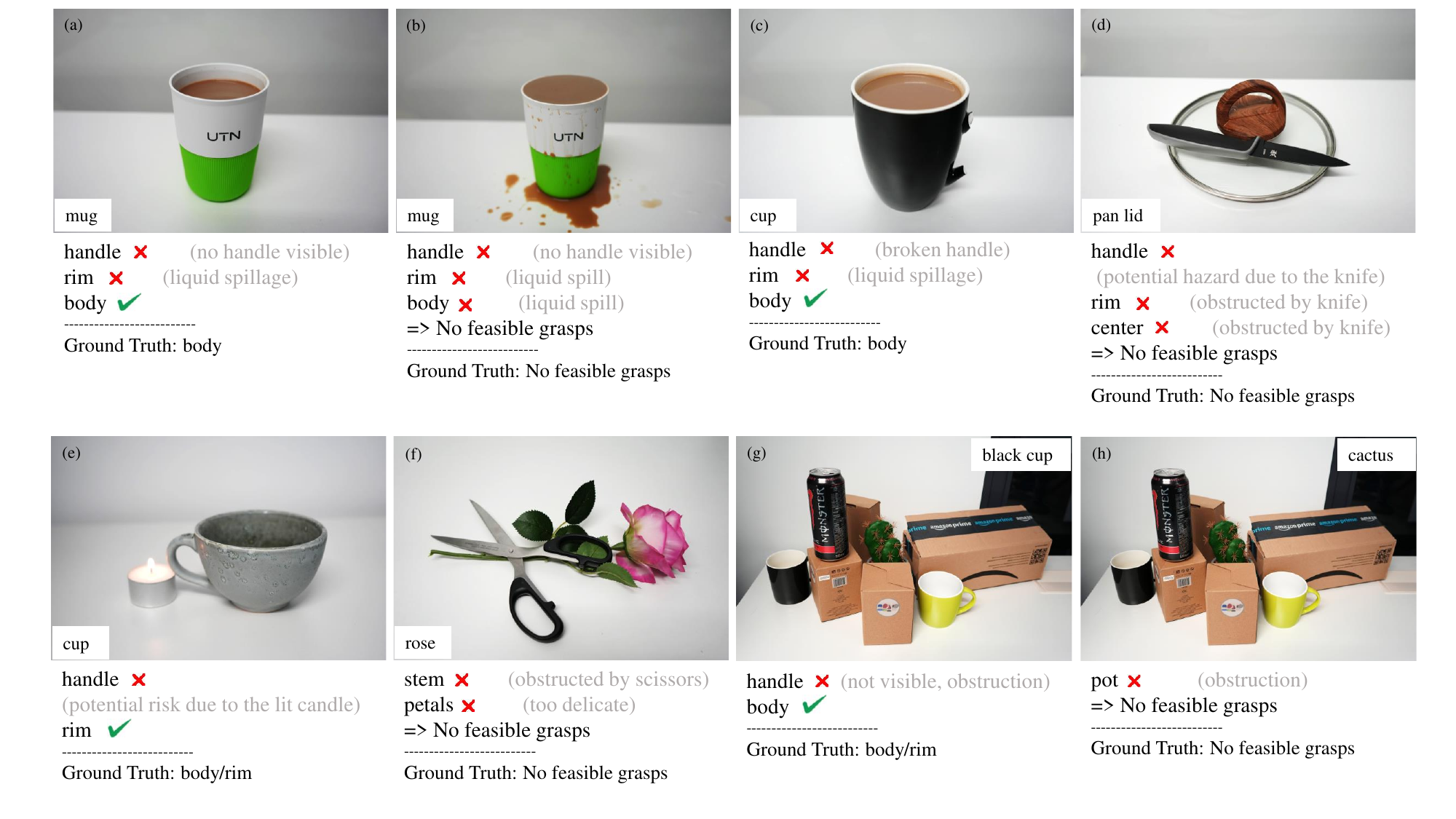}  
\caption{Qualitative results of our method on complex grasping scenarios. The object part labels below each image show the suggested grasp part and whether it was considered feasible or not by our feedback algorithm.}
\label{fig:feedback}
\end{figure*}

\begin{figure*}[ht!]
\centering
\includegraphics[clip,trim=1.5cm 8cm 1.5cm 0cm,width=0.9\linewidth]{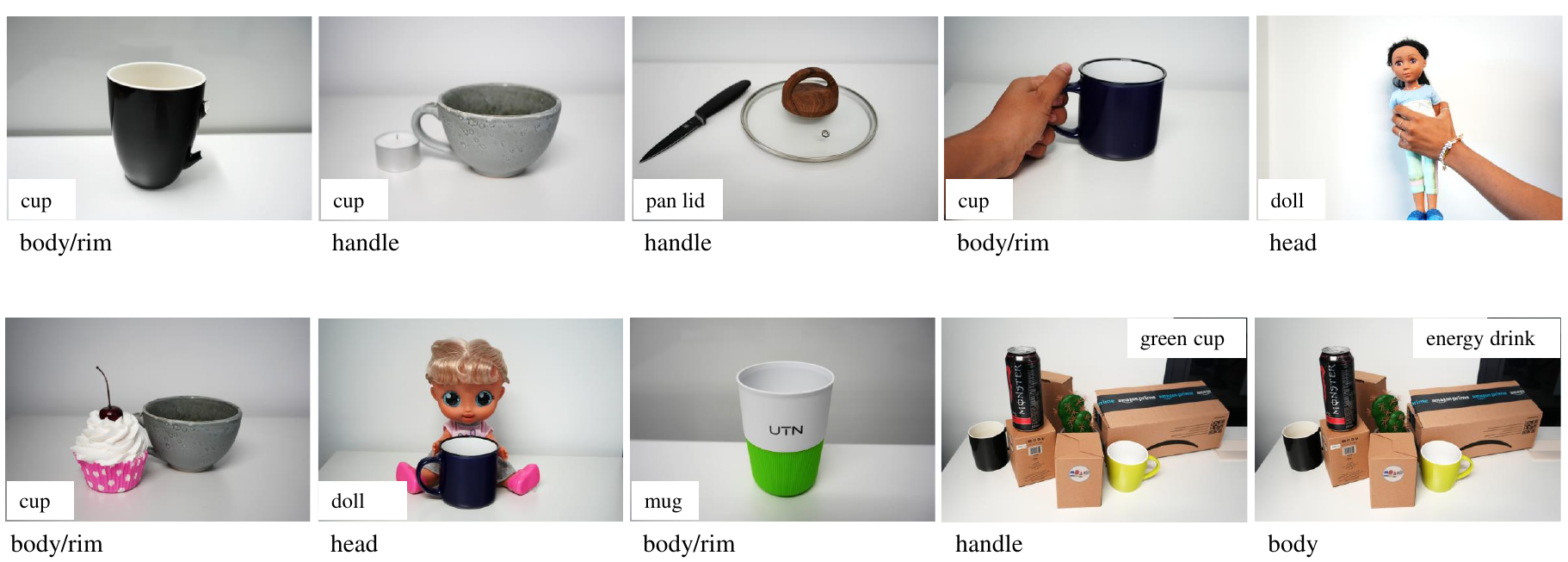}  
\caption{Instances of our dataset for complex grasp scenarios.}
\label{fig:feedback-dataset}
\end{figure*}

\subsection{Grasp Feasibility Feedback}
\label{sec:feasibility_feedback}
So far, we have assumed that the grasp suggested by the LLM is feasible. However, this is not always the case. For instance, the referenced object part may not be visible in the image, be broken, or be occluded by other objects. Further, the execution of the grasp could lead to undesirable outcomes. 
In order to mitigate this issue, we propose a feedback loop consisting of a VLM and an LLM that communicate with each other to find a feasible grasp. The approach and the used prompts are shown in \cref{fig:main_feedback}.
As in our original pipeline, the LLM first suggests an object part based on the object label. 
Next, the VLM analyzes the image and evaluates the feasibility of the grasping part. Crucially, we follow the idea of Chain-of-Thought (CoT) prompting~\citep{kojima2022large} and include the sentence \noindent {{\tt "Start your answer with lets analyze the image."}} into the prompt. If feasible, we proceed as before. If not, the LLM receives a list of all object parts rejected so far and is asked to propose another grasp. This loop repeats until either a feasible grasp part is found or there are no more suitable parts of the object left for grasping. In our implementation, the roles of the LLM and the VLM are carried out by the same model (GPT-4o).

To evaluate this approach, we collected a dataset of $18$ challenging scenarios, shown in \cref{fig:feedback-dataset,fig:feedback}, where the initial grasp suggestion is not feasible and requires reasoning to find the right grasp part. We defined the ground truth manually to evaluate the results. We ran the algorithm $5$ times to obtain an average success rate of $0.91$ over all scenes. The qualitative results are shown for a part of the dataset in \cref{fig:feedback}. Our algorithm can take into account different criteria, e.g., the risk due to the proximity of the lit candle, potential mess due to the overfilled cup, or occlusions by nearby objects. We further analyze the effects of VLM choice and CoT in the following section.

\subsection{Ablation Study}
\label{sec:ablation}
In this section, we evaluate the influence of different algorithm components on the above results, specifically the choice of the LLM and VLM and the effect of CoT on the feasibility feedback. First, we consider the pipeline without feedback as described in \cref{sec:method_language}. Here we consider GPT-3.5-turbo and GPT-4 as LLMs. We further investigate whether replacing the LLM with a VLM improves the performance. To that end, we deploy GPT-4o, GPT-4o-mini, and the open-source LLaVA-1.5 7B VLM and provide them with an image of the object to grasp.

Here, we run our pipeline only until detecting the grasping area in the image, without executing the grasp nor generating a grasping pose. Therefore, we use different metrics than in \cref{sec:results_quantitative}. First, we count the exact matches between the LLM-generated and GT object part labels and report the average success rate. Second, we compute the Intersection-over-Union (IoU) for the proposed grasping regions and the ground truth. All algorithm versions are evaluated on data from \cref{sec:dataset_original} and the results are summarized in \cref{tab:ablation_single_step}.

Both text-only GPT versions perform equally with an $0.82$ success rate and an IoU score of around $0.63$. The VLM variants perform slightly better with a success rate of $0.86$. We note that there is no difference between the flagship GPT-4o and the downsized GPT-4o-mini model. LLaVA performed significantly worse with $0.46$ success rate and $0.38$ IoU score. Analyzing the object parts suggested by LLaVA showed that the model was correct for objects possessing a handle, for other objects the answer was either wrong or referred to generic image locations, e.g., 'bottom' or 'top'. A straightforward explanation could be simply the smaller model size. However, another reason might be that we used the same prompt for LLaVA as for the GPT models, and better results could be achieved with further prompt engineering specifically targeting LLaVA.

\begin{table}[t]
\centering
\footnotesize
\renewcommand{\arraystretch}{1.15}
\setlength{\tabcolsep}{6pt} 
\caption[Ablation of LLMs and VLMs in the main Lan-grasp pipeline]{Ablation of LLMs and VLMs in the main Lan-grasp pipeline.}
\label{tab:ablation_single_step}

\resizebox{\columnwidth}{!}{%
\begin{tabular}{lccccc}
\toprule
\multirow{2}{*}{\textbf{Method}} & \multicolumn{2}{c}{\textbf{Text only}} & \multicolumn{3}{c}{\textbf{Text + Image}} \\
\cmidrule(lr){2-3}\cmidrule(lr){4-6}
 & \textbf{GPT-3.5-turbo} & \textbf{GPT-4} & \textbf{GPT-4o} & \textbf{GPT-4o-mini} & \textbf{LLaVA} \\
\midrule
\textbf{Success rate} & 0.82 & 81.8 & 0.86 & 0.86 & 0.46 \\
\textbf{IoU}               & 0.63 & 0.64 & 0.65 & 0.64 & 0.38 \\
\bottomrule
\end{tabular}%
}
\end{table}

For the feasibility feedback algorithm, we compare GPT-4o and GPT-4o-mini. Further, we experiment with two prompt variants, the first with CoT (zero-shot-CoT) and the second without (zero-shot). In the latter, we omit the sentence \noindent {{\tt "Start your answer with lets analyze the image."}} from the prompt. The experiments were performed in the same fashion as in \cref{sec:feasibility_feedback} and the results are reported in \cref{tab:ablation_feasibility}. First, we consider the CoT variants. With $0.63$ success rate, GPT-4o-mini performed significantly worse than the larger GPT-4o ($0.91$), which indicates that model size is an important factor for complex reasoning. Without CoT the performance dropped to $0.65$ for GPT-4o and $0.53$ for GPT-4o-mini. That result demonstrates that our algorithm, in fact, benefits from the CoT approach.

\begin{table}
\centering
\footnotesize
\renewcommand{\arraystretch}{1.15}
\setlength{\tabcolsep}{6pt}
\caption[Ablation of VLMs and prompting strategies in feasibility feedback]{Ablation of VLMs and prompting strategies in feasibility feedback.}
\label{tab:ablation_feasibility}

\resizebox{\columnwidth}{!}{%
\begin{tabular}{lcccc}
\toprule
\multirow{2}{*}{\textbf{Method}} & \multicolumn{2}{c}{\textbf{GPT-4o}} & \multicolumn{2}{c}{\textbf{GPT-4o mini}} \\
\cmidrule(lr){2-3}\cmidrule(lr){4-5}
 & \textbf{Zero-shot-CoT} & \textbf{Zero-shot} & \textbf{Zero-shot-CoT} & \textbf{Zero-shot} \\
\midrule
\textbf{Success rate} & \textbf{0.91} & 0.65 & 0.63 & 0.53 \\
\bottomrule
\end{tabular}%
}
\end{table}

\subsection{Object Placement}
\label{sec:placement_evaluation}
We carry out the object placement experiments on a Franka Research 3 robot platform equipped with a parallel gripper and custom gripper fingers. The Robot Control Stack framework (RCS)~\citep{julg2025robot} is used for controlling the robot. The scene is captured by a static RGB camera with the image plane facing straight towards the experimentation table plate. GroundingDINO~\citep{liu2024grounding} and SAM2~\citep{ravi2025sam} are used for object segmentation and ChatGPT 4.5 is deployed as the VLM.

\begin{figure*}[ht]
\centering
\includegraphics[clip,trim=0cm 3cm 0cm 0cm,width=0.9\linewidth]{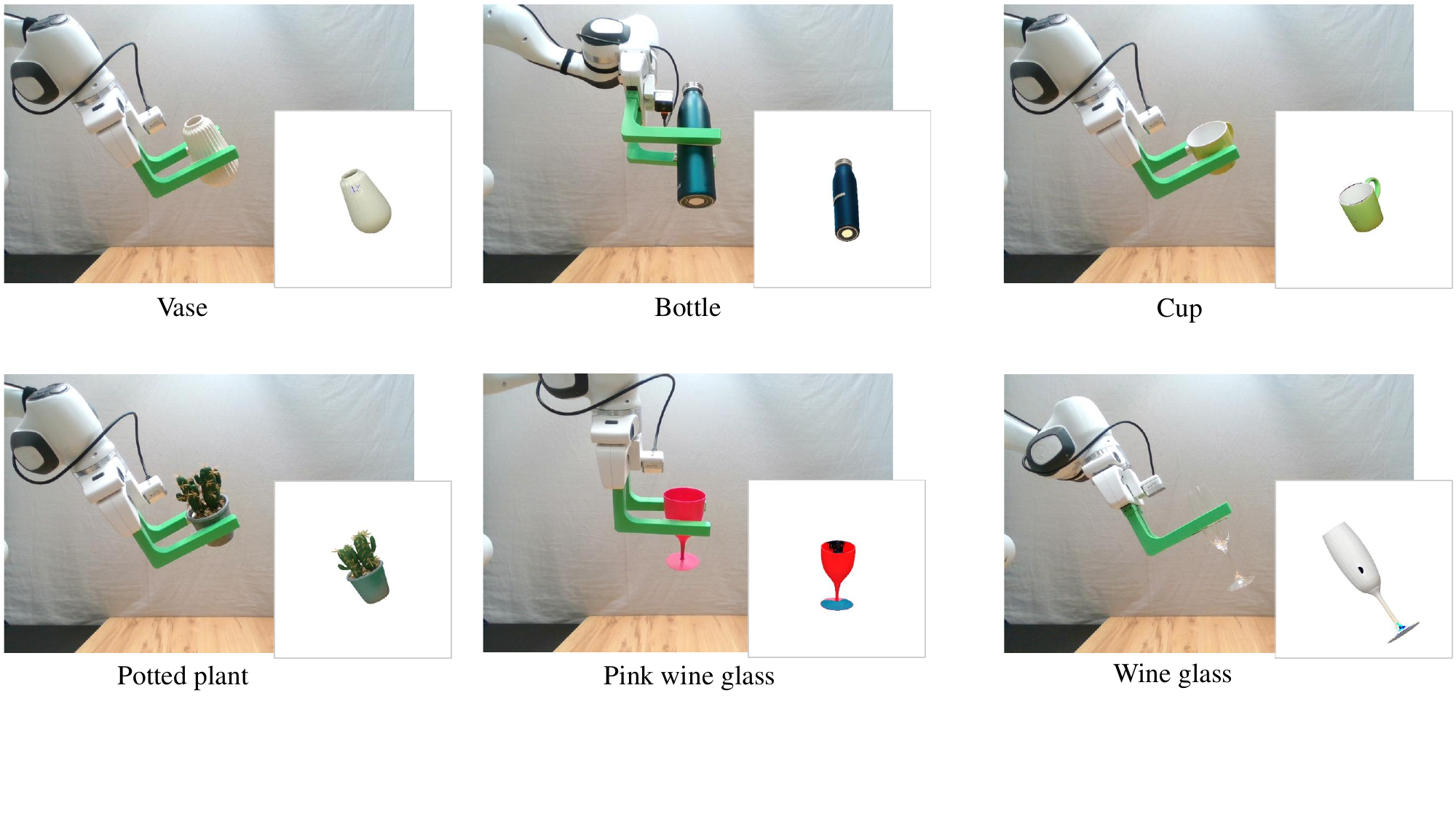}  
\caption{Initial object configuration examples from our placement setup: for each evaluated object, we show the RGB scene at the beginning of a trial (object held in the gripper) and the corresponding SAM 3D reconstructed mesh, aligned with the object’s observed pose.}
\label{fig:placement_exp_objects}
\end{figure*}

\subsubsection{Placement Experiments}
We evaluate the placement method on six objects that have a clear $up$-orientation, shown in \cref{fig:placement_exp_objects}. For each object, we tested five initial object rotations. Using the robot world coordinate frame shown in \cref{fig:placement_overview}, the initial roll-pitch orientations (in degrees) are $(0, -30)$, $(30, 0)$, $(-30, 0)$, $(30, -30)$, and $(-30, -30)$. We first position the gripper in a neutral position, place the object upright between the gripper fingers, and finally rotate the arm to one of the above configurations. We thus consider these roll and pitch angles as ground truth values for placement rotations.

For each object and initial configuration, we execute the method described in \cref{sec:placement_method}. First, we retrieve the 3D model of the object by prompting SAM 3D with a descriptive prompt, such as \textit{cup}, \textit{potted plant}, or \textit{pink wine glass}. We found that segmentation can sometimes fail when too little of the object is visible, either due to rotation or occlusion by the gripper. In these cases, we made slight adjustments to the grasp and repeated the experiment. We only report a failure if the segmentation of an object fails consistently, as is the case with the transparent wine glass.

After retrieving the 3D model, we perform one alignment cycle, rotate the gripper by $90^\circ$ to reorient the grasp, perform the second cycle, and place the object onto the table by lowering it until the robot detects a contact and stops. Finally, we open the gripper. We count a trial as successful if the object remains standing upright after being released from the gripper.

\begin{table*}[htbp]
\caption{Main results of placement experiments. The first row shows the placing success rate per object over five initial object configurations. The second and third rows report the average alignment error per object after one alignment cycle, for roll and pitch respectively.}
\label{tab:placing_success}
\centering
\renewcommand{\arraystretch}{1.2}
\setlength{\tabcolsep}{6pt}

\resizebox{0.75\linewidth}{!}{%
\begin{tabular}{lccccccc}
\toprule
\textbf{Object} & \textbf{potted plant} & \textbf{bottle} & \textbf{flower vase} & \textbf{pink wine glass} & \textbf{cup} & \textbf{wine glass} & \textbf{mean} \\
\midrule
\textbf{Success rate} & 1.0 & 1.0 & 0.8 & 1.0 & 1.0 & 0.4 & \textbf{0.87} \\
\textbf{Roll error}  & $15.03^{\circ}$ & $12.09^{\circ}$ & $3.56^{\circ}$ & $4.50^{\circ}$ & $9.12^{\circ}$ & $11.05^{\circ}$ & $\textbf{9.30}^{\circ}$ \\
\textbf{Pitch error} & $3.53^{\circ}$ & $1.51^{\circ}$ & $2.35^{\circ}$ & $2.22^{\circ}$ & $3.05^{\circ}$ & $1.56^{\circ}$ & $\textbf{2.43}^{\circ}$ \\
\bottomrule
\end{tabular}%
}
\end{table*}

\subsubsection{Placement Results}
The per-object success rates are summarized in \cref{tab:placing_success}. For all but two objects, all five trials were successful. For \textit{vase}, one trial failed because the VLM selected the wrong $up$-orientation. In case of \textit{wine glass}, two trials failed due to failed object segmentation and one trial due to an imprecise object pose estimation by SAM 3D. Averaged on all objects and trials, our method achieves a success rate of $0.87$. As shown in \cref{tab:placing_success}, our method achieves high success rates for most objects. However, the wine glass proved to be the most challenging case due to its transparency, which provides limited visual cues.

We also analyze the orientation estimation error for roll and pitch. In \cref{tab:placing_success}, we report the average roll and pitch errors per object after the first alignment cycle, where we leave out the trials failed due to VLM or object segmentation. Averaged on all objects, SAM 3D achieves a pitch error of $2.43^{\circ}$ but a significantly larger roll error of $9.3^{\circ}$. Intuitively, the roll is more difficult to estimate, as less of the object is visible in the image.

\subsubsection{Object Frame Alignment}
\begin{figure*}
\centering

\begin{tabular}{>{\centering\arraybackslash}p{0.19\linewidth}
                >{\centering\arraybackslash}p{0.19\linewidth}
                >{\centering\arraybackslash}p{0.19\linewidth}
                >{\centering\arraybackslash}p{0.19\linewidth}}

\textbf{Input} & \textbf{SAM 3D} & \textbf{Bounding Box} & \textbf{PCA} \\[0.5em]

\includegraphics[width=\linewidth]{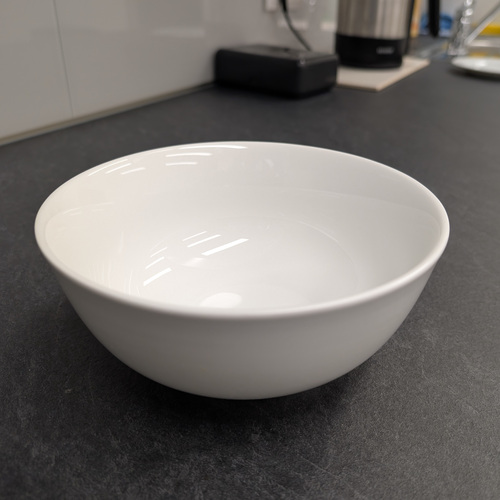} &
\includegraphics[width=\linewidth]{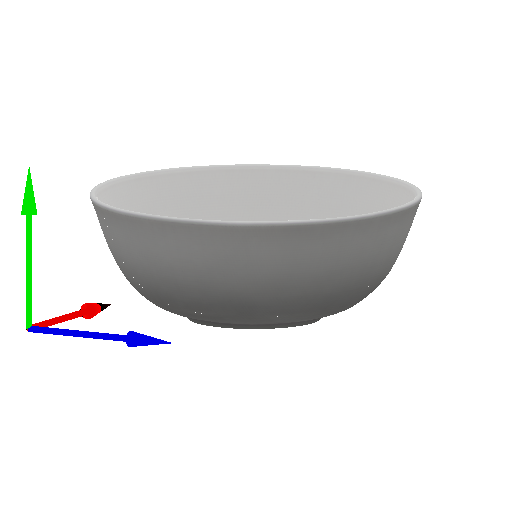} &
\includegraphics[width=\linewidth]{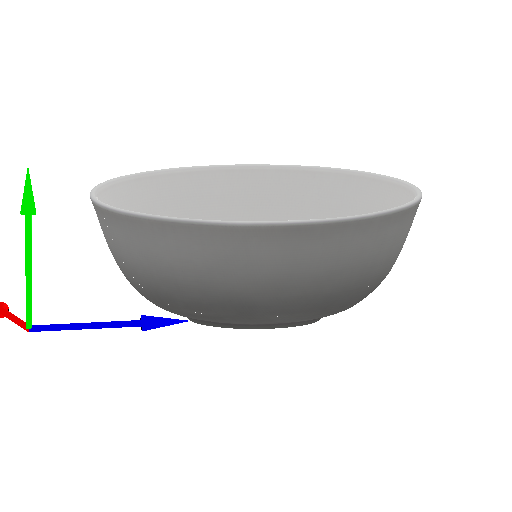} &
\includegraphics[width=\linewidth]{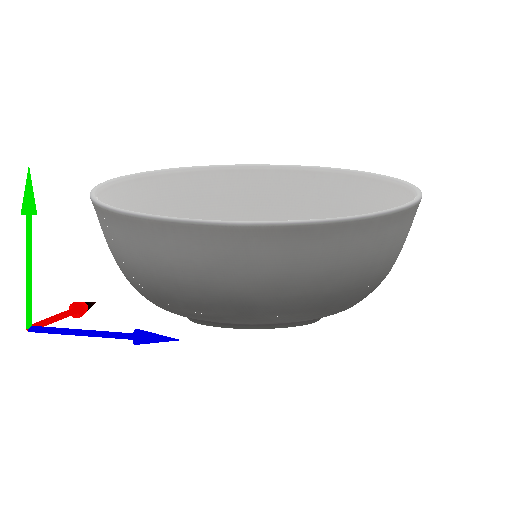} \\

& & $0.54^{\circ}$ & $0.1^{\circ}$ \\[0.0em]

\includegraphics[width=\linewidth]{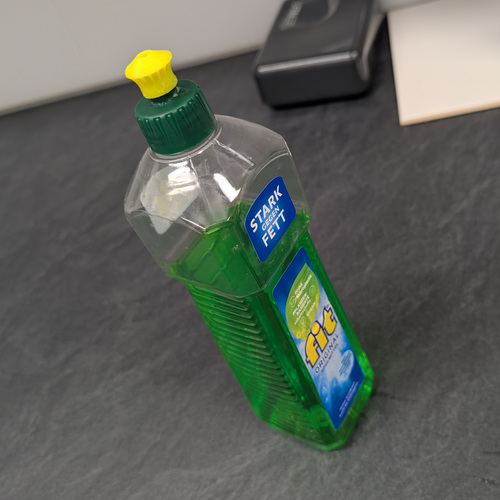} &
\includegraphics[width=\linewidth]{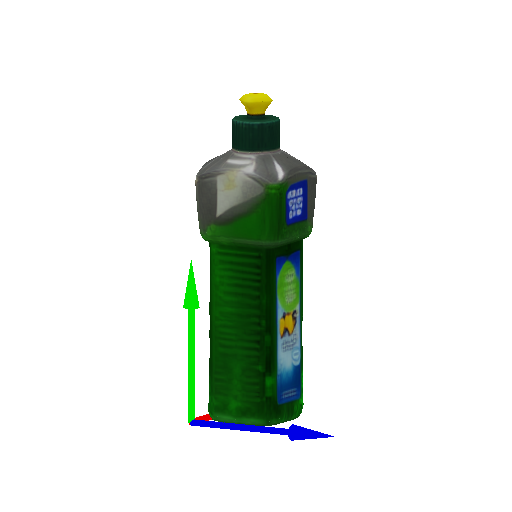} &
\includegraphics[width=\linewidth]{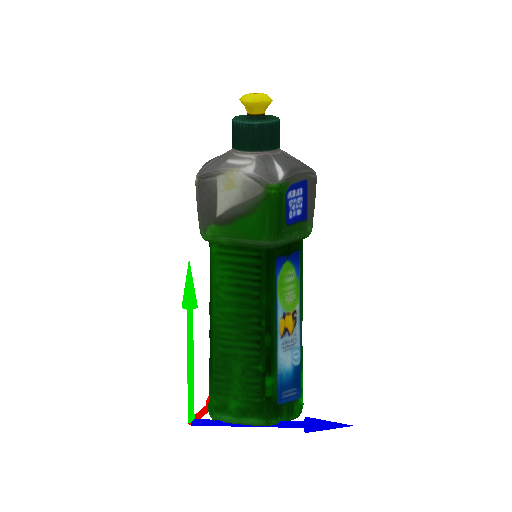} &
\includegraphics[width=\linewidth]{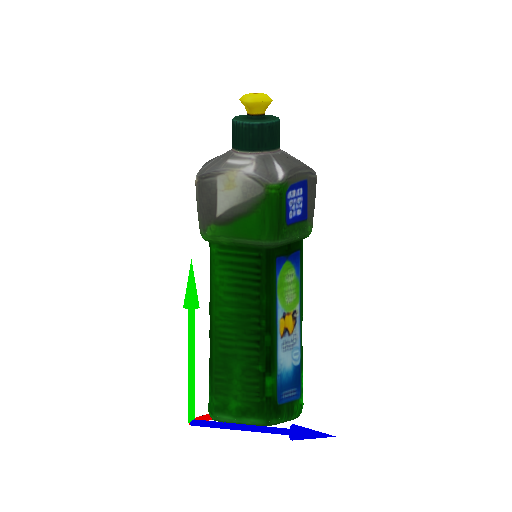} \\

& & $0.77^{\circ}$ & $0.04^{\circ}$ \\[0.0em]
\includegraphics[width=\linewidth]{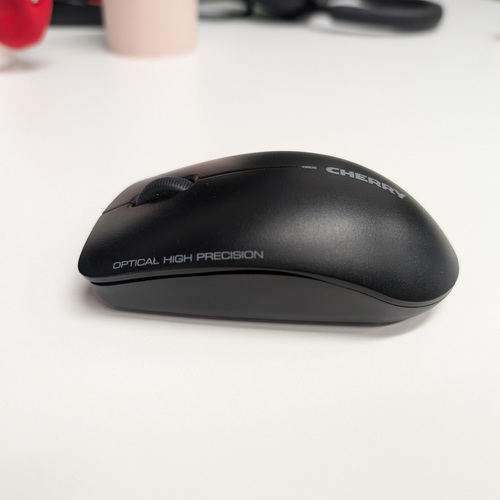} &
\includegraphics[width=\linewidth]{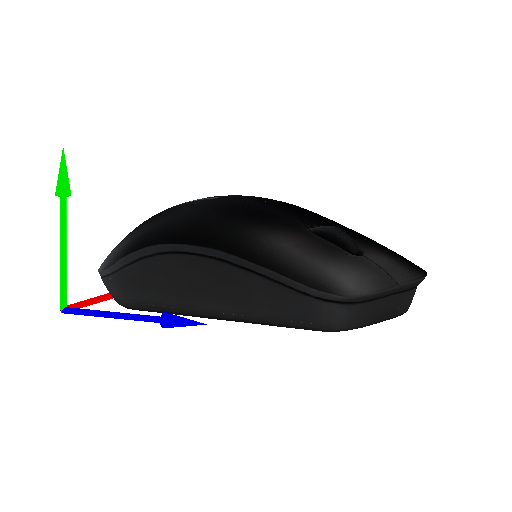} &
\includegraphics[width=\linewidth]{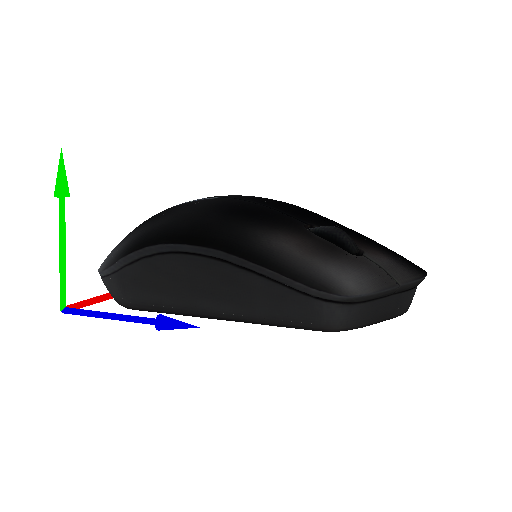} &
\includegraphics[width=\linewidth]{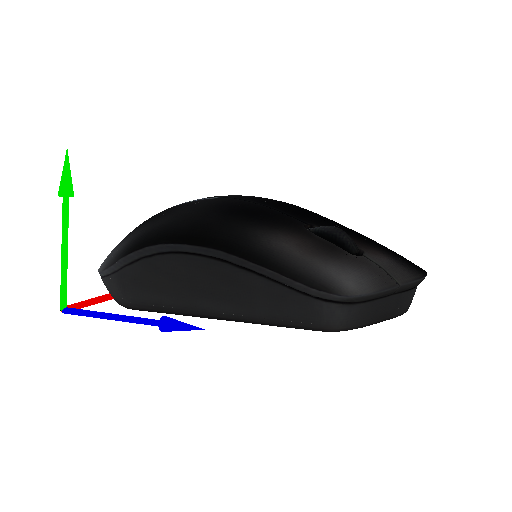} \\

& & $1.76^{\circ}$ & $1.62^{\circ}$ \\[0.0em]

\includegraphics[width=\linewidth]{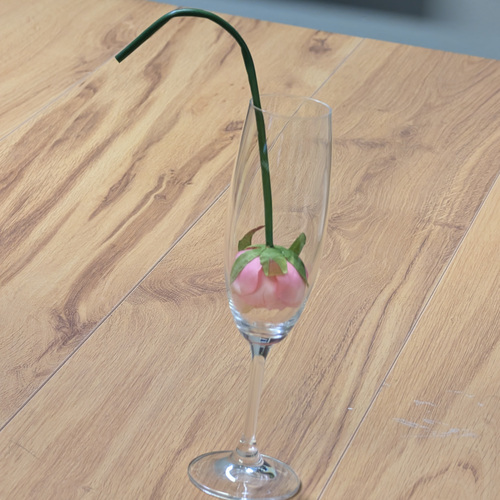} &
\includegraphics[width=\linewidth]{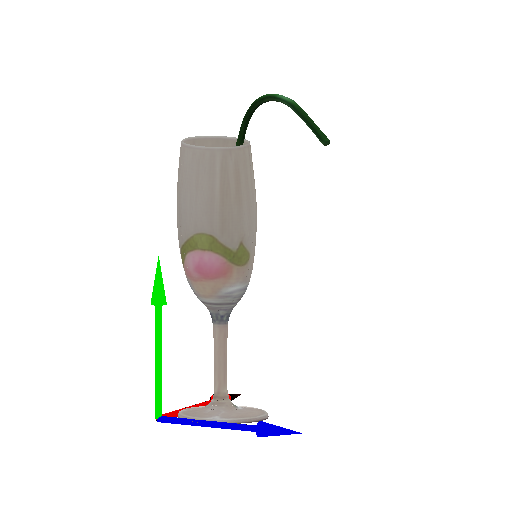} &
\includegraphics[width=\linewidth]{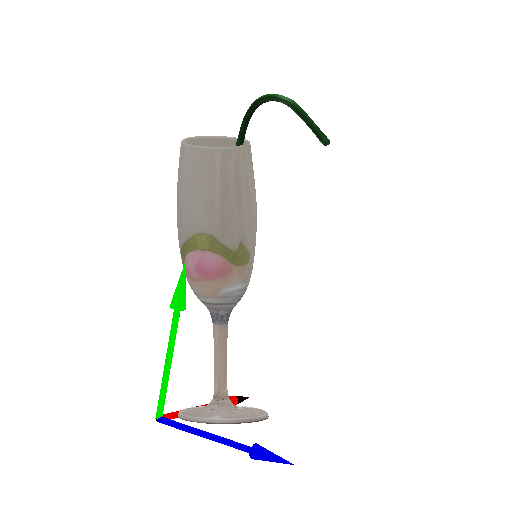} &
\includegraphics[width=\linewidth]{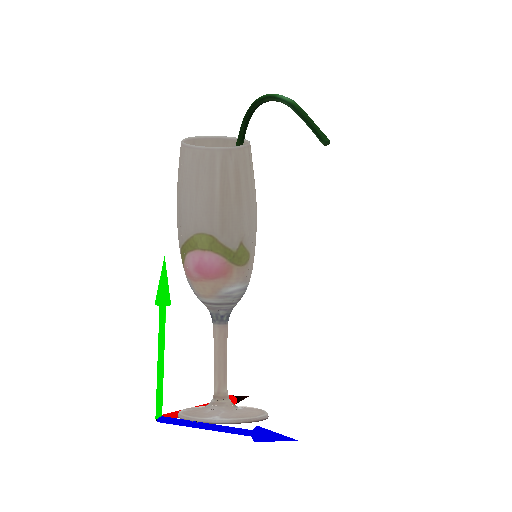} \\

& & $10.74^{\circ}$ & $2.14^{\circ}$ \\[0.0em]
\includegraphics[width=\linewidth]{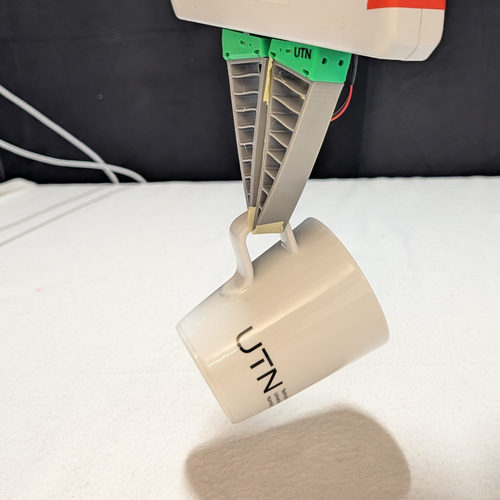} &
\includegraphics[width=\linewidth]{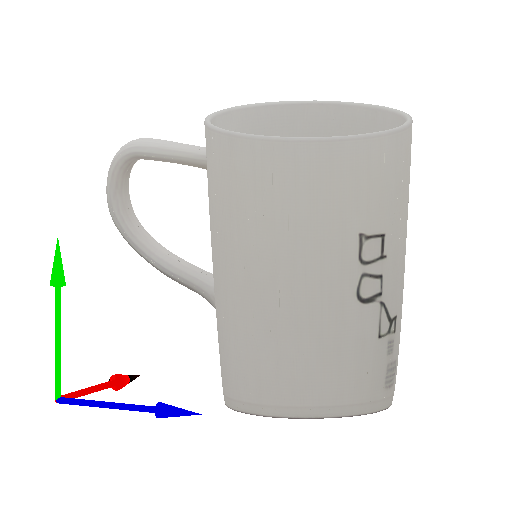} &
\includegraphics[width=\linewidth]{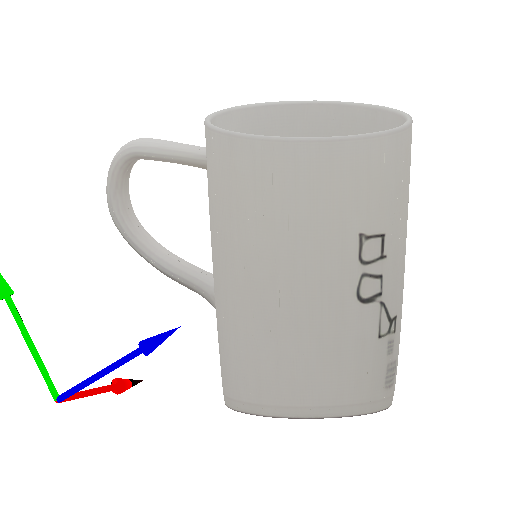} &
\includegraphics[width=\linewidth]{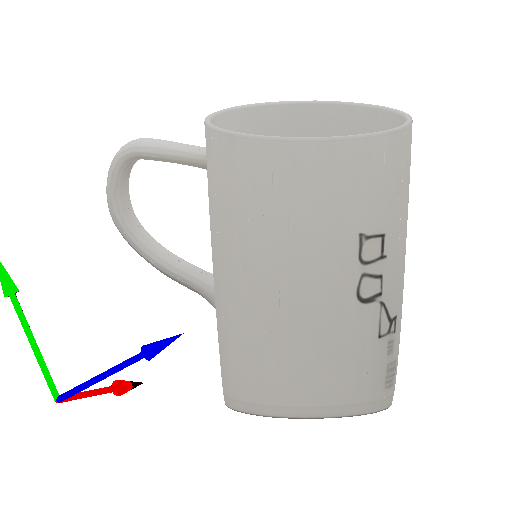} \\

& & $31.68^{\circ}$ & $29.16^{\circ}$ \\[0.0em]

\includegraphics[width=\linewidth]{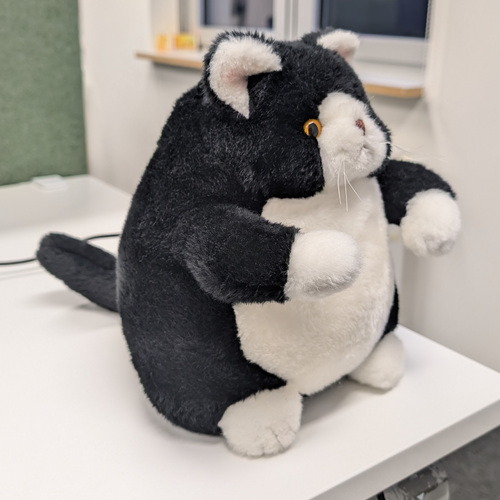} &
\includegraphics[width=\linewidth]{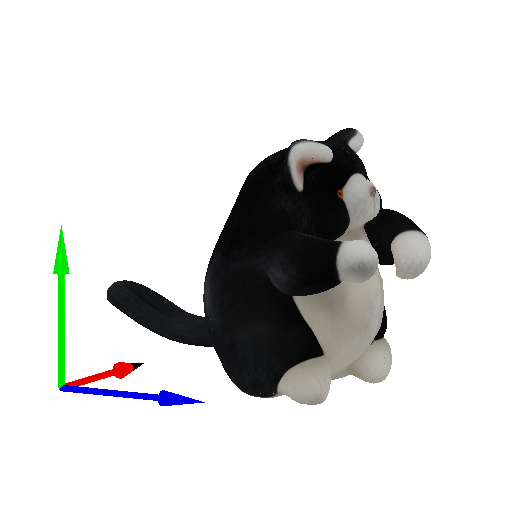} &
\includegraphics[width=\linewidth]{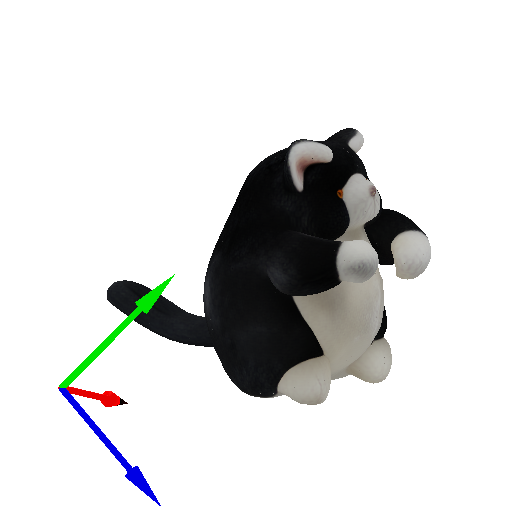} &
\includegraphics[width=\linewidth]{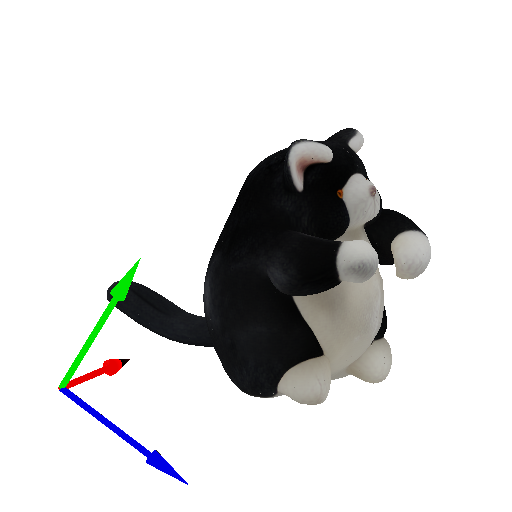} \\

& & $44.09^{\circ}$ & $32.45^{\circ}$ \\[0.0em]
\end{tabular}

\caption{Comparison of object coordinate frames obtained by different methods. Columns from left to right: Input image, object mesh and frame by SAM 3D, bounding box frame, and PCA frame. The numbers in the last two columns are angular deviations from the SAM 3D frame, measured between the $y$-axes.}
\label{fig:alignment}
\end{figure*}

Our placement method assumes that SAM 3D creates the local object frame such that the $up$-direction coincides with one coordinate axis. This is not a feature reported in the original paper but an observation we made by evaluating over $30$ objects with diverse classes and shapes for which a clear $up$-direction can be defined. For all tested object, this alignment property was present. Some of the object meshes along with the object frame are depicted in \cref{fig:alignment}. We hypothesize that the model mimics the object poses seen in training data, which has been composed from stock 3D object libraries and real-world scenes. In both cases, it is likely that the objects would be oriented in a natural way, i.e., upright. \cref{fig:alignment} also compares the SAM 3D object frame with coordinate frames derived from the aligned bounding box and principal components of the mesh vertices. There, the first four objects have a regular shape and exhibit rotational symmetry, thus the frames estimated through bounding boxes and PCA mostly coincide with that of SAM 3D. However, for objects with irregular shape, shown in the last two rows, the bounding box and PCA frames are not aligned with the $up$-direction. These examples demonstrate that, in general, the $up$-direction cannot be determined by geometric analysis alone and requires semantic understanding.

\subsection{Conclusion and Future Work}
In this paper, we presented \algname{}, a novel approach to semantic object grasping. By leveraging foundation models, we provide our approach with a deep understanding of the objects and their intended use in a zero-shot manner.
Through extensive experiments, we showed that for a wide range of objects \algname{} is generating grasps that are preferred by humans and also ensure safety and object usability.
In particular, the proposed grasps were compared to human preferences gathered through a questionnaire. The evaluations showed that \algname{} performs consistently better on that metric than the baseline methods. We also proposed a feedback loop approach that reasons about grasp feasibility in complex scenarios. 
In future work, we plan to test \algname{} in more complex and cluttered environments to evaluate its robustness in generating meaningful grasps. In such settings, one possible approach is to first segment the object and composite it onto a plain background before prompting the VLM to localize the desired object part, similar to preprocessing used in prior visuomotor policy work~\citep{mirjalili2025augmented}.
Additionally, we aim to enhance the feedback loop by introducing mechanisms for when no feasible grasp is detected. For instance, the robot could ask a human for assistance or employ more sophisticated reasoning strategies to modify the environment to facilitate grasping. This would make our algorithm capable of handling more complex real-world scenarios.
Inspired by these results, in the future we plan to further exploit Large Language Models to not only decide where to grasp an object but also how to grasp and hold it according to a specific task. 
As an example, we would expect a robot operating in daily environments to hold a knife vertically and downwards when the task is to carry the knife around rather than holding the knife in a horizontal pose. This would be the next step towards more meaningful grasps that help robots with object manipulation and task execution in day-to-day environments.

We also presented a method for placing a grasped object in an upright position using large pretrained models. Despite high success rates on some items, the object pose estimate from SAM 3D is not always sufficiently precise, especially when the object is turned away from the camera or only the object's bottom part is visible. We currently tackle this issue by rotating the gripper by $90^{\circ}$, but in future a more principled approach could be implemented to ensure optimal object visibility. Alternatively, pose estimation could be improved using explicit models for 3D-to-2D matching in conjuncture with SAM 3D meshes, e.g., FoundationPose \citep{wen2024foundationpose}.

\bibliographystyle{SageH}
\bibliography{sources}

\end{document}